\documentclass[english,authoryear]{elsarticle} 

\usepackage{lineno,hyperref}
\usepackage{booktabs} 

\usepackage{array}
\newcolumntype{L}[1]{>{\raggedright\let\newline\\\arraybackslash\hspace{0pt}}m{#1}}
\newcolumntype{C}[1]{>{\centering\let\newline\\\arraybackslash\hspace{0pt}}m{#1}}
\newcolumntype{R}[1]{>{\raggedleft\let\newline\\\arraybackslash\hspace{0pt}}m{#1}}

\usepackage{babel} 
\usepackage{apalike}

\usepackage{amsmath}

\makeatletter
\def\ps@pprintTitle{%
 \let\@oddhead\@empty
 \let\@evenhead\@empty
 \def\@oddfoot{}%
 \let\@evenfoot\@oddfoot}
\makeatother

\begin{document}
	
	\begin{frontmatter}
		
		\title{Automatic segmenting teeth in X-ray images: Trends, a novel data set, benchmarking and future perspectives}
		
		\author{Gil Silva\fnref{fn1}}
		\fntext[fn1]{gil.jader@gmail.com}
		
		\author{Luciano Oliveira\fnref{fn2}}
		\fntext[fn2]{lrebouca@ufba.br}
		
		\address{Ivision Lab, Federal University of Bahia, Brazil}
		
		\author{Matheus Pithon\fnref{fn3}}
		\fntext[fn3]{matheuspithon@gmail.com}
		
		\address{Southeast State University of Bahia, Brazil}
		
	\begin{abstract}
			
			This review presents an in-depth study of the literature on segmentation methods applied in dental imaging. Ten segmentation methods were studied and categorized according to the type of the segmentation method (region-based, threshold-based, cluster-based, boundary-based or watershed-based), type of X-ray images used (intra-oral or extra-oral) and characteristics of the dataset used to evaluate the methods in the state-of-the-art works. We found that the literature has primarily focused on threshold-based segmentation methods (54\%). 80\% of the reviewed papers have used intra-oral X-ray images in their experiments, demonstrating preference to perform segmentation on images of already isolated parts of the teeth, rather than using extra-oral X-rays, which show tooth structure of the mouth and bones of the face. To fill a scientific gap in the field, a novel data set based on extra-oral X-ray images are proposed here. A statistical comparison of the results found with the 10 image segmentation methods over our proposed data set comprised of 1,500 images is also carried out, providing a more comprehensive source of performance assessment. Discussion on limitations of the methods conceived over the past year as well as future perspectives on exploiting learning-based segmentation methods to improve performance are also provided.\\					
	\end{abstract}
		
	\begin{keyword}	
		\texttt{image segmentation} \sep dental X-ray  \sep orthopantomography
	\end{keyword}
		
	\end{frontmatter}  
	
	\section{Introduction}
	
	In dentistry, radiographic images are fundamental data sources to aid diagnosis. Radiography is the photographic record of an image produced by the passage of an X-ray source through an object \citep{Quinn1980}. X-ray images are used in dental medicine to check the condition of the teeth, gums, jaws and bone structure of a mouth \citep{Quinn1980}. Without X-rays, Dentists would not be able to detect many dental problems until they become severe. This way, the radiographic examination helps the dentist to discover the cause of the problem at an early stage, allowing then to outline the best treatment plan for the patient. Another application of dental X-rays is in the field of forensic identification, especially in cadavers \citep{Paewinsky2005}. The forensic dentistry aims to identify individuals based on their dental characteristics. In recent years, forensic literature has also provided automatic methods to assessing person's age from degenerative changes in teeth \citep{Willems2002}. These age-related changes can be assessed by digital radiography \citep{Paewinsky2005}. With the advancement of artificial intelligence and pattern recognition algorithms, X-ray images have been increasingly used as an input to these intelligent algorithms. In this context, we highlight here an in-depth study over some segmentation methods in the literature that are regarded to the recognition of image patterns in dental X-rays.
	
	\subsection{Overview of dental image segmentation}
	
	In dentistry, X-rays are divided into two categories: (i) Intra-oral radiographic examinations are techniques performed with the film positioned in the buccal cavity (the X-ray image is obtained inside the patient's mouth); and (ii) extra-oral radiographic examinations are the techniques in which the patient is positioned between the radiographic film and the source of X-rays (the X-ray image is obtained outside the patient's mouth) \citep{AmericanDentalAssociation1987}.
	
	In this paper, some works that use segmentation methods applied to the following types of X-ray images are analyzed: bitewing and periapical (intra-oral), and panoramic (extra-oral). The \textbf{bitewing} X-ray images are used to show details of the upper and lower teeth in a mouth region, while the \textbf{periapical} X-ray images is used to monitor the entire tooth \citep{Wang2016}. On the other hand, \textbf{panoramic} radiography, also known as orthopantomography, is one of the radiological exams capable of obtaining fundamental information for the diagnosis of anomalies in dental medicine \citep{Amer2015}, \citep{Wang2016}. Orthopantomographic examination allows for the visualization of dental irregularities, such as: teeth included, bone abnormalities, cysts, tumors, cancers, infections, post-accident fractures, temporomandibular joint disorders that cause pain in the ear, face, neck and head \citep{Oliveira2011}.
	
	X-ray images are pervasively used by dentists to analyze the dental structure and to define patient's treatment plan. However, due to the lack of adequate automated resources to aid the analysis of dental X-ray images, X-ray analysis relies on mostly the dentist's experience and visual perception \citep{Wang2016}. Other details in dental X-rays that make it difficult to analyze these images are: Variations of patient-to-patient teeth, artifacts used for restorations and prostheses, poor image qualities caused by certain conditions (such as noise, low contrast, homogeneity in regions close to objects of interest), space existing by a missing tooth, and limitation of acquisition methods; all these challenges result in unsuccessful development of automated computer tools to aid dental diagnosis, avoiding completely automatic analysis \cite{Amer2015}.
	
	Image segmentation is the process of partitioning a digital image into multiple regions (pixel set) or objects, in order to make an image representation simpler, and to facilitate its analysis. The present work is being carried out to help in finding advances in the state-of-the art of methods for segmenting dental X-ray images that are able, for example, to isolate teeth from other parts of the image (jaws, temporomandibular regions, details of nasal, face and gums) towards facilitating the automatic analysis of X-rays. With that, we are capable to discuss limitations in the current proposed methods and future perspectives for breakthroughs in this research field.
	
	\subsection{Contributions}
	
	This paper provides an in-depth review of the literature in dental X-ray image segmentation. A comparative evaluation of ten methods to segment extra-oral dental images over a novel data set is also addressed. The proposed data set was gathered specially for the present study, and contains 1,500 annotated panoramic X-ray images \footnote{Our data set will be publicly available on the acceptance of the paper.}. This present study is towards to answering the following questions (see Section \ref{sec_taxonomy}): Which category of segmentation method is most used in the reviewed works?, do public data sets used to evaluate dental segmentation methods present sufficiently variability to evaluate the progress of the field?. Also, these other following questions (see Section \ref{sec_seg}) which segmentation method obtains the best performance on extracting characteristics of radiographic images (panoramic X-ray), so that it is possible to perfectly isolate teeth?, what are the gaps in dental X-rays that can benefit from the application of image segmentation methods? Finally, we discuss recent advances in pattern recognition methods that could be applied in tooth segmentation (see Section \ref{sec:discussion}). 
    
    To answer the list of questions, the present review follows the steps: (i) analysis of the current state-of-the-art, observing the trends of the segmentation methods in dental X-ray images, (ii) identification of which image segmentation methods are the most used among the reviewed works, (iii) analysis of the amount and variety of images used in the experiments of the reviewed works, (iv) identification of which type of dental X-ray image has been most used among the reviewed works, (v) introduction of a novel annotated data set with a high variability and a great number of images, and, finally, (vi) a comparative evaluation of dental segmentation methods applied to our data set. These steps are followed from the classification of the papers found in the literature, considering: segmentation methods, X-ray image types, size and variety of the data sets used. 
	
	It is noteworthy that the reviewed articles mostly work with small data sets, ranging from 1 to 100 images in average, and the only work with more than one thousand images is not publicly available, or only containing images varying only in relation to the number of teeth. To tackle this limitation, the proposed data set is comprised of 1,500 annotated images, which allow the classification of X-rays in 10 categories according to the following general characteristics: Structural variations in relation to the teeth, number of teeth, existence of restorations, existence of dental implants, existence of dental appliances, and existence of dental images with more than 32 teeth. The images represent the most diverse situations found among patients in dental offices. In this sense, a comparative evaluation of 10 segmentation methods was performed to verify which method can more accurately identify each individual tooth, in panoramic X-ray images. Metrics, such as accuracy, specificity, precision, recall (sensitivity) and F-score, were used to assess the performance of each segmentation method analyzed here.
	
	\section{Research methodology} \label{sec_methodology}
	
	This review has followed the methodological steps: (A) select the digital libraries and articles (Section \ref{sec_sources}), (B) review the selected articles (Section \ref{sec_review}), (C) define relevant categories to classify the articles and classify articles in the categories defined (Section \ref{sec_taxonomy}). Steps (B) and (C) ran until final results were obtained. Step (D) was repeated until the evaluation of all the segmentation methods studied was finalized (Section \ref{sec_seg}). Finally, step (E) presents discussion about the evaluated methods and future directions to build more robust and efficient segmentation methods (Section \ref{sec:discussion}). 
	
	
	\subsection{Research sources and selection of the articles} \label{sec_sources}
	
	Our review is based on the state-of-the-art articles found in the following digital libraries: IEEE Xplore\footnote [1]{http://ieeexplore.ieee.org/Xplore}, ScienceDirect\footnote [2]{http://www.sciencedirect.com}, Google Scholar\footnote [3]{https://scholar.google.com} and Scopus\footnote [4]{http://www.scopus.com/}. The choice of these four digital libraries relies on the fact that they include articles presented in all other digital libraries related to either Computer Science or Dentistry. The selection of the articles was based on their prominence in the field of English language. The articles were selected in two phases: In phase I, a total of 94 articles were found in these four digital libraries. In Phase II, articles such as calendars, book chapter, publisher's notes, subject index, volume content, and from symposiums were excluded from the present study. \textbf{Only peer-reviewed international conferences and journal articles were considered}; among those, studies that corresponded to some of the following cases were considered as non-relevant and excluded from the analysis: (1) did not answer any of our questions in this research, (2) duplicated, (3) not peer-reviewed, and (4) did not apply segmentation methods on at least one of the following types of dental X-rays: Bitewing, periapical or panoramic. The final number of articles selected was reduced to 41, at the end. The number of articles initially found in each digital library is summarized in Table \ref{tab:Total_number_of_studies}. 
	
	As shown in Table \ref{tab:Total_number_of_studies}, only three of the four libraries surveyed have found relevant studies. In addition, forty-nine percent (49\%) of the articles selected as relevant were found in the Science Direct digital library. Table \ref{tab:Dist_articles} shows the initial statistics obtained in the review stage of the present study, containing the distribution of articles by digital library and year of publication. The data presented in Table \ref{tab:Dist_articles} show the largest number of articles found in the IEEE Xplore digital library with 34 articles (36\%). The results in Table \ref{tab:Dist_articles} also show the increasing trend in the number of articles published in recent years. Sixty-six percent (66\%) of the articles found were published in the last five years (61 articles). 
	
	\subsection{Selection of the relevant articles} \label{sec_review}
	
	In the second stage, the goal was to ensure the correct classification of the articles selected only as relevant. The review of the articles follows a categorization phase (presented in the next section), since it was necessary to re-read articles to classify them in each of the respective categories. 
	
	\begin{table} [!ht] \footnotesize 
		\centering
		\caption{Total number of studies found by digital library.}
		\begin{tabular}{lC{1.2cm}C{1.4cm}C{1.5cm}C{1.8cm}C{1.4cm}}
			\toprule
			\textbf{SOURCE}
			& \textbf{Source Results}
			& \textbf{Not Relevant}
			& \textbf{Repeated}
			& \textbf{Incomplete}
			& \textbf{Relevant Studies}
			\\ \midrule
			\textbf{IEEE Xplore}    & 34 & 16 & 0 & 5 & 13
			\\ \textbf{Science Direct}  &   28  & 8 &   0  & 0 & 20
			\\ \textbf{Google Scholar}  &  23   & 7 &  1   & 7 & 8
			\\ \textbf{Scopus} &  9   & 0 &  1   & 8 & 0
			\\ \midrule
			\textbf{TOTAL} &  \textbf{94}   & 31 &   2  & 20 & \textbf{41}
			\\ \bottomrule
		\end{tabular}
		\label{tab:Total_number_of_studies}
	\end{table}
	
	\begin{table}[!ht] \footnotesize
		\centering
		\caption{Distribution of articles by digital library and year of publication.}
		\begin{tabular}{C{1.1cm}ccC{1.8cm}C{1.9cm}C{2.0cm}C{0.9cm}}
			\toprule
			\textbf{YEAR}
			& \textbf{Sum}
			& \%
			& IEEE Xplore
			& Science Direct
			& Google Scholar
			& Scopus
			\\ \midrule
			\textbf{2016} & 11 & 12\% & 3 & 4 & 2 & 2 
			\\ \textbf{2015} & 12 & 13\% & 5 & 6 & 1 & 0 
			\\ \textbf{2014} & 12 & 13\% & 5 & 2 & 3 & 2 
			\\ \textbf{2013} & 13 & 14\% & 8 & 1 & 1 & 3 
			\\ \textbf{2012} & 13 & 14\% & 5 & 4 & 4 & 0 
			\\ \textbf{2011} & 5 & 5\% & 2 & 0 & 3 & 0 
			\\ \textbf{2010} & 6 & 6\% & 1 & 3 & 2 & 0 
			\\ \textbf{2009} & 5 & 5\% & 1 & 1 & 2 & 1 
			\\ \textbf{2008} & 5 & 5\% & 1 & 1 & 2 & 1 
			\\ \textbf{2007} & 8 & 9\% & 1 & 4 & 3 & 0 
			\\ \textbf{2006} & 3 & 3\% & 2 & 1 & 0 & 0 
			\\ \textbf{2005} & 0 & 0\% & 0 & 0 & 0 & 0 
			\\ \textbf{2004} & 1 & 1\% & 0 & 1 & 0 & 0 
			\\ \midrule
			\textbf{Total} & 94 & 100\% & 34 & 28 & 23 & 9 
			\\ \bottomrule
		\end{tabular}			
		\label{tab:Dist_articles}
	\end{table}
		
	\section{Taxonomy of the relevant works} \label{sec_taxonomy}
	
	Each article selected as relevant was classified among categories defined in the present study, according to: The segmentation method used, type of dental X-ray used images, the size and variety of the data set used. It is noteworthy that the segmentation methods discussed and benchmarked in this review are strictly from the state-of-the-art works.
	
	\subsection{Segmentation categories} \label{sec_review_segmentation_methods}
	
	\cite{Chen2004} categorize segmentation methods according to the characteristics (shape, histogram, threshold, region, entropy, spatial correlation of pixels, among others) searched in a variety of source images (X-ray, thermal ultrasonic, etc) to generate the cut-off point (value that determines what the objects of interest in the analyzed image are). We adapted the general classification found in \citep{Chen2004} to the classification of the works studied in the field of dental image segmentation as follows: (1) Region-based, (2) threshold-based, (3) cluster-based, (4) boundary-based, (5) watershed-based. The categories were defined based on the characteristics that the relevant articles explore in the images analyzed to carry out the segmentation. Table \ref{tab:Works_grouped_by_segmentation_methods} shows the relevant works, classified into the categories of the segmentation methods. Important details about each segmentation method presented in each category are addressed in Section \ref{sec_Segmentation_methods_parsed}.
	
	\begin{table}[!ht] \footnotesize
		\centering
		\caption{Works grouped by segmentation methods.}
		\begin{tabular}{p{3.4cm}|p{7.8cm}}
			\toprule
			\textbf{Category} & \textbf{Segmentation method} (\textbf{Related works}) \\
			\midrule
			Region-based & Region growing (\citep{Lurie2012}, \citep{Modi2011}) 
			\\
			\\
			Threshold-based & Histogram-based threshold (\citep{Abaza2009}, \citep{Ajaz2013} \citep{Cameriere2015}, \citep{Jain2004}, \citep{Lin2014}, \citep{Dighe2012}, \citep{Huang2008}, \citep{Lin2015}, \citep{Bruellmann2016}, \citep{Amer2015}, \citep{Tikhe2016}) / Variable threshold (\citep{Said2006},  \citep{Geraets2007}, \citep{Lin2010}, \citep{Wang2016}, \citep{Nomir2008a}, \citep{Kaur2016}, \citep{Nomir2008}, \citep{Keshtkar2007}, \citep{Lin2013}, \citep{Indraswari2015}, \citep{MohamedRazali2014}) 
			\\
			\\
			Cluster-based & Fuzzy-C-means (\cite{Alsmadi2015}, \cite{Son2016}) 
			\\
			\\
			Boundary-based  & Level set method (\citep{EhsaniRad2013}, \citep{Li2006}, \citep{Li2007}, \citep{An2012}) / Active contour (\citep{Ali2015}, \citep{Niroshika2013}, \citep{Hassan2016}) / Edge detection (\citep{Senthilkumaran2012}, \citep{Lin2012}, \citep{Razali2015}, \citep{Senthilkumaran2012a}, \citep{Grafova2013}, \citep{Trivedi2015}) / Point detection (\citep{Economopoulos2008}) 
			\\	
			\\
			Watershed-based & Watershed (\citep{Li2012}) 
			\\	
			\bottomrule
		\end{tabular}
		\label{tab:Works_grouped_by_segmentation_methods}
	\end{table}
	
	\paragraph{\textbf{Region-based}}
	
	The goal of the region-based method is to divide an image into regions, based on discontinuities in pixel intensity levels. Among the relevant articles selected, only \cite{Lurie2012} and \cite{Modi2011} used the region-based segmentation. The aim of the study in \citep{Lurie2012} was to segment panoramic X-ray images of the teeth to assist the Dentist in procedures for detection of osteopenia and osteoporosis. \cite{Modi2011} used region growing approach to segment bitewing X-ray images. 
	
	\paragraph{\textbf{Threshold-based}} The rationale of the intensity threshold application in image segmentation starts from the choice of a threshold value. Pixels whose values exceed the threshold are placed into a region, while pixels with values below the threshold are placed into an adjacent region. Most of the articles selected as relevant (54\%) use the threshold-based segmentation approach \citep{Abaza2009}, \citep{Ajaz2013} \citep{Cameriere2015}, \citep{Jain2004}, \citep{Lin2014}, \citep{Dighe2012}, \citep{Huang2008}, \citep{Lin2015}, \citep{Bruellmann2016}, \citep{Amer2015}, \citep{Tikhe2016}), (\citep{Said2006},  \citep{Geraets2007}, \citep{Lin2010}, \citep{Wang2016}, \citep{Nomir2008a}, \citep{Kaur2016}, \citep{Nomir2008}, \citep{Keshtkar2007}, \citep{Lin2013}, \citep{Indraswari2015}, \citep{MohamedRazali2014}.
	
	In certain cases, pixel gray levels, which belongs to the objects of interest, are substantially different from the gray levels of the pixels in the background. In those cases, threshold segmentation based on the histogram of the image is usually used to separate objects of interest from the background. This way, histograms can be used in situations, where objects and background have intensity levels grouped into two dominant modes. The present research identified that seven out of the relevant papers used \textbf{histogram-based threshold} as the main stage of segmentation \citep{Abaza2009}, \citep{Ajaz2013} \citep{Cameriere2015}, \citep{Jain2004}, \citep{Lin2014}, \citep{Dighe2012}, \citep{Huang2008}, \citep{Lin2015}, \citep{Bruellmann2016}, \citep{Amer2015}, \citep{Tikhe2016}.
	
	Thresholding simply based on the histogram of the image usually fails when the image exhibits considerable variation in contrast and illumination, resulting in many pixels that can not be easily classified as first or second plane. One solution to this problem is to try to estimate a "shading function", and then use it to compensate for the pattern of non-uniform intensities. The commonly used approach to compensate for irregularities, or when there is a lot of variation of the intensity of the pixels related to the dominant object (in which case the histogram-based thresholding has difficulties) is the use of variable threshold based on local statistics of the pixels of the image. The studies in \citep{Said2006},  \citep{Geraets2007}, \citep{Lin2010}, \citep{Wang2016}, \citep{Nomir2008a}, \citep{Kaur2016}, \citep{Nomir2008}, \citep{Keshtkar2007}, \citep{Lin2013}, \citep{Indraswari2015}, \citep{MohamedRazali2014} applied \textbf{local variable thresholding} as the main step for segmentation of the dental X-ray images.
	
	\paragraph{\textbf{Cluster-based}}
	
	Clustering is a method used to make automatic grouping of data according to a certain degree of similarity between the data. The criterion of similarity depends on the problem to be solved. In general, the number of groups to be detected must be informed as the initial parameter for the algorithm to perform data clustering. Among the relevant papers, \cite{Alsmadi2015} used clustering to perform the segmentation of panoramic X-ray images, while \cite{Son2016} proposed a clustering-based method to segment X-rays of bitewing and periapical types.
	
	\paragraph{\textbf{Boundary-based}}
	
	Boundary-based methods are used to search for discontinuities (point and edge detection) in the gray levels of the image. Thirty-four percent (34\%) of the relevant papers used boundary-based segmentation methods.
	
	The classical boundary-based approach performs the search for points and edges in images by detecting discontinuity in color or pixel intensities in images. Among the works that used boundary-based methods, \citep{Senthilkumaran2012}, \citep{Lin2012}, \citep{Razali2015}, \citep{Senthilkumaran2012a}, \citep{Grafova2013}, \citep{Trivedi2015} and \citep{Economopoulos2008} used the classical approach for point and edge detection to segment the images. A more recent approach on boundary-based segmentation is known as active contour model \citep{Ali2015}, \citep{Niroshika2013}, \citep{Hassan2016}, also called snakes, which performs segmentation by delineating an object outline from an image. The goal is to minimize the initialization of energy functions, and the stop criterion is when the minimum energy is detected. The region that represents the minimum energy value corresponds to the contour that best approaches the perimeter of an object. Another recent boundary-based detection approach is a variation of the active contour model known as level set method (LSM). The LSM performs segmentation by means of geometric operations to detect contours with topology changes. The studies found in \citep{EhsaniRad2013}, \citep{Li2006}, \citep{Li2007}, \citep{An2012} used LSM to segment the X-ray images.
	
	\paragraph{\textbf{Watershed-based}}
	
	Watershed is a transformation defined in a grayscale image. The watershed transformation uses mathematical morphology to segment an image in adjacent regions. Among the relevant articles selected, only \citep{Li2012} used the watershed-based segmentation to segment bitewing X-ray images.
	
	\subsection{Type of the X-ray images} 
	
	Approximately eighty percent (80\%) of the reviewed papers used intra-oral X-ray images. Only three of the reviewed papers used extra-oral panoramic X-ray images. The studies addressed in \citep{Geraets2007}, \citep{Son2016} and \citep{Trivedi2015} perform experiments with intra-oral and extra-oral images. Table \ref{tab:Works_grouped_by_X-ray} summarizes the relevant papers grouped by the type of X-ray image. 
	
	\begin{table}[!ht] \footnotesize 
		\centering
		\caption{Works grouped by X-ray images.}
		\begin{tabular}{l|p{6.7cm}}
			\toprule
			\textbf{X-ray} & \textbf{Related works} \\
			\midrule
			Bitewing   &  \citep{Jain2004}, \citep{EhsaniRad2013}, \citep{Senthilkumaran2012}, \citep{Nomir2008}, \citep{Lin2010}, \citep{Lin2012}, \citep{Wang2016}, \citep{Keshtkar2007}, \citep{Nomir2008a}, \citep{Modi2011}, \citep{Ali2015}, \citep{Li2012}, \citep{Kaur2016} \\\\
			Periapical  & \cite{Cameriere2015}, \citep{Lin2014}, \citep{Li2006}, \citep{Dighe2012}, \citep{Li2007}, \citep{Huang2008}, \citep{Lin2015}, \citep{Bruellmann2016}, \citep{Lin2013}, \citep{Niroshika2013}, \citep{Tikhe2016}, \citep{Senthilkumaran2012a}, \citep{An2012}, \citep{Economopoulos2008} \\\\
			Panoramic  & \citep{Alsmadi2015}, \citep{Amer2015}, \citep{Lurie2012}, \citep{Ajaz2013}, \citep{Indraswari2015}, \citep{MohamedRazali2014}, \citep{Razali2015}, \citep{Hassan2016}, \citep{Grafova2013} \\\\
			Bitewing / Periapical & \citep{Said2006}, \citep{Abaza2009} \\\\
			Bitewing / Panoramic & \citep{Son2016} \\\\
			Periapical / Panoramic  & \citep{Geraets2007} \\\\
			Bitewing / Periapical / Panoramic & \citep{Trivedi2015} \\
			\bottomrule
		\end{tabular}
		\label{tab:Works_grouped_by_X-ray}
	\end{table}
	
	\subsection{Characteristics of the data sets used in the reviewed works} \label{sec:datasets}
	
    \begin{figure}[!ht]
		\begin{center}
			\includegraphics[scale=0.48]{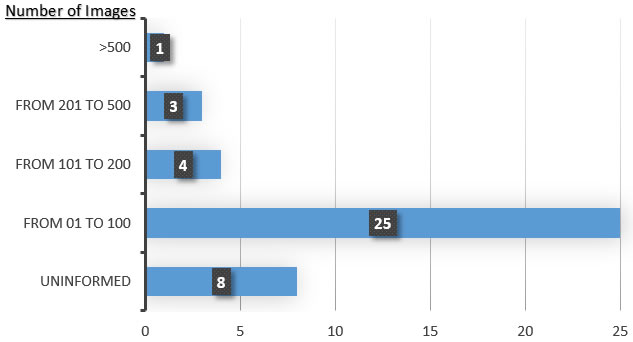}
			\caption{Number of works by number of images} \label{figura:number_images}
		\end{center}
	\end{figure}
    
	Sixty-one percent (61\%) of the relevant papers used data sets containing between 1 and 100 X-ray images (\citep{Lin2012}, \citep{Li2007}, \citep{Son2016}, \citep{Alsmadi2015}, \citep{Lin2015}, \citep{Lin2010}, \citep{Cameriere2015}, \citep{Jain2004}, \citep{Dighe2012}, \citep{Lin2014}, \citep{Bruellmann2016}, \citep{Economopoulos2008}, \citep{Grafova2013}, \citep{Kaur2016}, \citep{An2012}, \citep{EhsaniRad2013}, \citep{Tikhe2016}, \citep{Lin2013}, \citep{Ajaz2013}, \citep{MohamedRazali2014}, \citep{Indraswari2015}, \citep{Modi2011}, \citep{Amer2015}, \citep{Li2012},  \citep{Senthilkumaran2012a}). Eight of the reviewed articles did not present information about the data set used (\citep{Li2006}, \citep{Senthilkumaran2012}, \citep{Trivedi2015}, \citep{Niroshika2013}, \citep{Ali2015}, \citep{Razali2015}, \citep{Keshtkar2007}, \citep{Geraets2007}). Four of the papers reviewed used between 101 and 200 images (\citep{Wang2016}, \citep{Nomir2008a}, \citep{Lurie2012}, \citep{Nomir2008}). Three used between 201 and 500 images (\citep{Huang2008}, \citep{Abaza2009}, \citep{Hassan2016}). Only one among the papers reviewed used more than 500 images in their experiments \citep{Said2006}. In general, the reviewed articles exploited data sets containing X-ray images with small variations (\textit{i.e.}, varying only with respect to the number of teeth). In addition, as shown in the previous section, there is a predominance of intra-oral radiographs (which show only a part of the teeth) rather than extra-oral radiographs (which present the entire dental structure in a single image). Figure \ref{figura:number_images} depicts the number of works versus number of images used in each group of work.
	
	\section{Evaluation of the segmentation methods} \label{sec_seg}
	
	In our work, to evaluate the segmentation methods studied, we created a methodology that consists of six stages. In the \textbf{first} stage, we started with the acquisition of images through the orthopantomograph (device used for the generation of orthopantomography images), and the collected images were classified into 10 categories according to the variety of structural characteristics of the teeth. The \textbf{second} stage consists of annotating the images (obtaining the binary images), which correspond to the demarcations of the objects of interest in each analyzed image. After finishing the tooth annotation process, in the \textbf{third} stage, the buccal region is annotated, as the region of interest (ROI) to determine the actual image of the teeth. In the \textbf{fourth} stage, the statistics of the gathered data set are calculated. The \textbf{fifth} and \textbf{sixth} consist in analyzing the performance of the segmentation algorithms, using the metrics summarized in Table \ref{tab:metrics}, and in evaluating the results achieved by each segmentation method studied. 
	
	\subsection{Construction of the data set} 
	
	The images used in our data set were acquired from the X-ray camera model: ORTHOPHOS XG 5 / XG 5 DS / Ceph, manufactured by Sirona Dental Systems GmbH. X-rays were acquired at the Diagnostic Imaging Center of the Southwest State University of Bahia (UESB). The radiographic images used for this research were coded in order to identify the patient in the study\footnote{The use of the radiographs in the research was authorized by the National Commission for Research Ethics (CONEP) and by the Research Ethics Committee (CEP), under the report number 646,050, approved on 05/13/2014.}.
	
	The gathered data set consists of 1,500 annotated panoramic X-ray images. The images have significant structural variations in relation to: the teeth, the number of teeth, existence of restorations, existence of implants, existence of appliances, existence of supernumerary teeth (referring to patients with more than 32 teeth), and the size of the mouth and jaws. All images originally obtained by the ORTHOPHOS XG 5 / XG 5 DS / Ceph ortopantomograph had dimensions 2440 $\times$ 1292 pixels. The images were captured in gray level. The work with panoramic X-ray images is more challenging, due to heterogeneity reasons, among which the following stands out: 1) Different levels of noise generated by the ortopantomograph; 2) Image of the vertebral column, which covers the front teeth in some cases; 3) Low contrast, making morphological properties complex.
			
\begin{figure}[!ht]
		\begin{center}
			\includegraphics[scale=0.4]{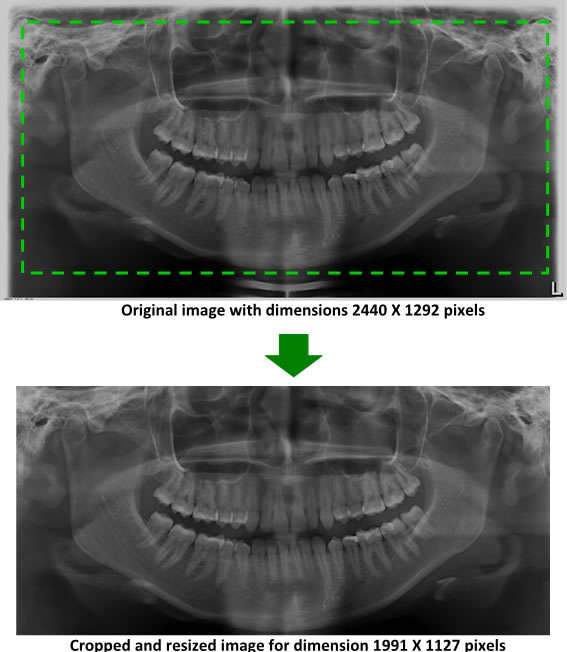}
			\caption{Example of the clipping and resizing of the data set images of the present work.} \label{figura:recorte_imagem}
		\end{center}
	\end{figure}

\begin{figure}[!ht]
		\begin{center}
			\includegraphics[scale=0.6]{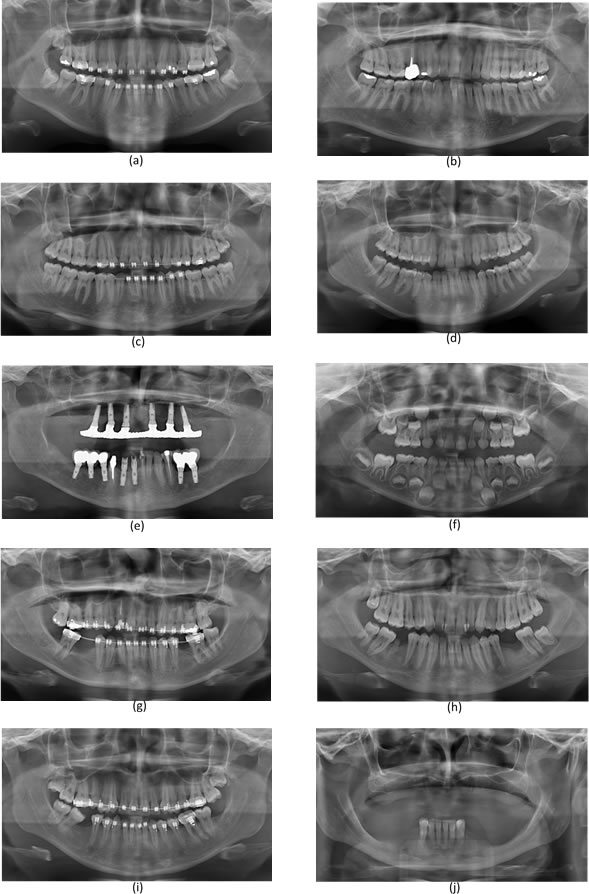}
			\caption{Examples of images from the data set categories of present work: \textbf{(a) Category 1}; \textbf{(b) Category 2}; \textbf{(c) Category 3}; \textbf{(d) Category 4}; \textbf{(e) Category 5}; \textbf{(f) Category 6}; \textbf{(g) Category 7}; \textbf{(h) Category 8}; \textbf{(i) Category 9}; \textbf{(j) Category 10}.} \label{figura:imagens_dataset}
		\end{center}
	\end{figure}

	To thoroughly benchmark the methods studied here, the 1,500 images were distributed among 10 categories. The images were named, using whole numbers, in sequential order by category, aiming at not identifying the patients in the study. The process of categorizing the images was performed manually, selecting images individually, counting tooth by tooth, as well as verifying structural characteristics of the teeth. The images were classified according to the variety of structural characteristics of the teeth (see Table \ref{tab:statistics}). Finally, the images were cut out to disregard non-relevant information (white border around the images and part of the spine) generated by the orthopantomograph device. After the clipping process, there was a change in the size of the images to 1991 $\times$ 1127 pixels, but without affecting the objects of interest (teeth), as shown in Figure \ref{figura:recorte_imagem}. The cropped images were saved on the new dimension to be used in the following stages, which will be presented in the next sections. Figure \ref{figura:imagens_dataset} shows an X-ray image corresponding to each of the categories of our data set.
	
	\begin{figure}[!ht]
		\begin{center}
			\includegraphics[scale=0.55]{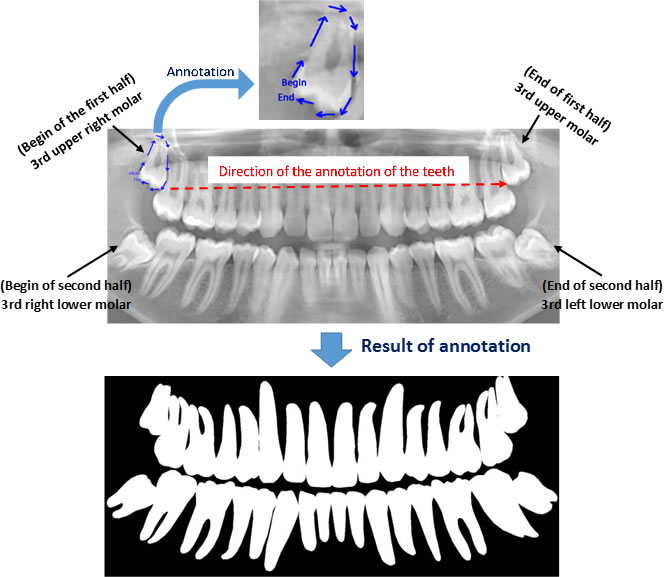}
			\caption{Annotation of the teeth.} \label{figura:processo_anotacaoA}
		\end{center}
	\end{figure} 
	
	\begin{figure}[!ht]
		\begin{center}
			\includegraphics[scale=0.56]{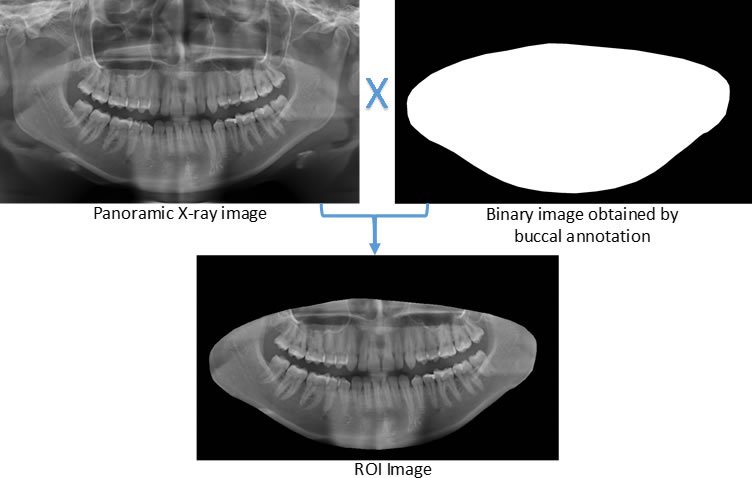}
			\caption{Determining the ROI of the images.} \label{figura:anotacao_boca}
		\end{center}
	\end{figure}
	
	\paragraph{\textbf{Image annotation}}
	
	The process of annotating the images of our proposed data set occurred in two parts. First, it was initiated by the upper jaw through the annotation of the third right upper molar and making the annotation of all the teeth of the upper arch to the third left upper molar. Then, the same process was performed on the lower jaw with all the teeth, and in the same direction as the upper jaw, from left to right, starting with the annotation of the third right lower molar, and annotating all teeth from the lower arch to the lower third molar. Figure \ref{figura:processo_anotacaoA} illustrates the tooth annotation process through a panoramic X-ray image of the data set. 
	
	\paragraph{\textbf{Determining ROI}}
	
	For each image, after the annotation of the teeth, the buccal region was also annotated, covering the whole region delineated by the contour of the jaws. This process was carried out in view of preserving the area containing all the teeth (objects of interest). Finally, the region of interest (ROI) was determined by multiplying the values of the pixel array elements, representing the original panoramic X-ray image, by its corresponding binary matrix, resulting from the process of oral annotation. Figure \ref{figura:anotacao_boca} illustrates the whole process to determine the ROI of the images.
	
	\begin{table}[!ht] \footnotesize
		\centering
		\caption{Categorization of data set images and average number of teeth per category.}
		\begin{tabular}{c | p{6.8cm} | c | C{1.3cm}}
			\toprule
			\textbf{Number} & \textbf{Category} & \textbf{Images} & \textbf{Average teeth} \\
			\midrule
			1   &  Images with \textbf{all the teeth}, containing teeth with restoration and with dental appliance & 73 & 32  \\\hline
			
			2   &  Images with \textbf{all the teeth}, containing teeth with restoration and without dental appliance & 220 & 32  \\\hline
			
			3   &  Images with \textbf{all the teeth}, containing teeth without restoration and with dental appliance & 45 & 32  \\\hline
			
			4   &  Images with \textbf{all the teeth}, containing teeth without restoration and without dental appliance & 140 & 32  \\\hline
			
			5   &  Images containing dental implant & 120 & 18  \\\hline
			
			6   &  Images containing more than 32 teeth & 170 & 37  \\\hline
			
			7   &  Images \textbf{missing teeth}, containing teeth with restoration and dental appliance & 115 & 27  \\\hline
			
			8   &  Images \textbf{missing teeth}, containing teeth with restoration and without dental appliance & 457 & 29  \\\hline
			
			9   &  Images \textbf{missing teeth}, containing teeth without restoration and with dental appliance & 45 & 28  \\\hline
			
			10   &  Images \textbf{missing teeth}, containing teeth without restoration and without dental appliance & 115 & 28  \\
			\bottomrule
		\end{tabular}
		\label{tab:statistics}
	\end{table}
	
	\paragraph{\textbf{Data set statistics}}
	
	Table \ref{tab:statistics} presents the statistics of our data set: The categorization of the images, the total number of images used, the total of images by category and the average of teeth of the images by category.
	
	\begin{table}[!ht] \footnotesize
		\centering
		\caption{Image statistics by category.}
		\begin{tabular}{ccccc}
			\toprule
			\multicolumn{1}{c}{\textbf{Category}} & \textbf{Highest value} & \textbf{Lowest value} & \multicolumn{1}{c}{\textbf{Mean}} &  \multicolumn{1}{c}{\textbf{Entropy}} \\
			\midrule
			Category 1 & 253   & 10    & 108.30 & 6.93 \\
			Category 2 & 250   & 16    & 108.29 & 6.83 \\
			Category 3 & 248   & 13    & 107.25 & 6.88 \\
			Category 4 & 215   & 20    & 107.31 & 6.82 \\
			Category 5 & 254   & 5     & 109.36 & 6.94 \\
			Category 6 & 230   & 18    & 100.43 & 6.86 \\
			Category 7 & 255   & 7     & 108.50 & 6.88 \\
			Category 8 & 253   & 11    & 106.72 & 6.89 \\
			Category 9 & 251   & 9     & 107.33  & 6.89 \\
			Category 10 & 214   & 20    & 105.94 & 6.70 \\
			\bottomrule
		\end{tabular}
		\label{tab:Overall_averages_of_categories}
	\end{table}
	
	\paragraph{\textbf{Statistics of the image ROIs}}
	
	For all statistics, only the pixels in the image ROIs were considered. The results of the statistical operations were used as a parameter to perform the segmentation algorithms studied. The statistics raised over the image ROIs were as follows:
    
	The image statistics per category were organized in a single table to better analyze the results found among the categories, as shown in Table \ref{tab:Overall_averages_of_categories}. From the analysis of that table, it was possible to compare the characteristics of each data set category. For instance, \textbf{category 5} is formed by images with dental implants, which correspond to regions of high luminosity in the images, resulting in pixels with greater intensity than those found in the images of the other categories.
	
	\begin{table}[!ht] \footnotesize
		\centering
		\caption{Metrics used to evaluate the segmentation methods studied.}
		\begin{tabular}{p{2.7cm}|p{8.4cm}}
			\toprule
			\multicolumn{2}{c}{\textbf{Initial measures}} \\
			
			\midrule
			Positive (P) & Pixel is in a class of interest \\
			\hline
			Negative (N) & Pixel is not in a class of interest \\
			\hline
			True Positive (TP) & The pixel in the ground truth is positive, while method ranks the pixel as positive \\
			\hline
			True Negative (TN) & The pixel in; 3)  the ground truth is negative, while method ranks the pixel as negative \\
			\hline
			False Positive (FP) & The pixel in the ground truth is negative, while method ranks the pixel as positive \\
			\hline
			False Negative (FN) & The pixel in the ground truth is positive, while method ranks the pixel as negative \\
			\midrule
			
			\multicolumn{2}{c}{\textbf{Metrics used for performance evaluation}}  \\
			\midrule
			
			\textit{\textbf{Accuracy}} & Relation between total of hits on the total set of errors and hits. This value is calculated by: (TP + TN)/(TP + FN + FP + TN) \\
			\hline
			\textit{\textbf{Specificity}} & Percentage of negative samples correctly identified on total negative samples. This value is calculated by: TN/(FP + TN) \\
			\hline
			\textit{\textbf{Precision}} & Percentage of positive samples correctly classified on the total of samples classified as positive. This value is calculated by: TP/(TP + FP) \\
			\hline
			\textit{\textbf{Sensitivity/Recall}} & Percentage of positive samples correctly classified on the total of positive samples. This value is calculated by: TP/(TP + FN) \\
			\hline
			\textit{\textbf{F-score}} & Represents the harmonic average between precision and sensitivity. It is calculated by: 2*Recall*Precision/(Recall + Precision) \\		
			\bottomrule
		\end{tabular}
		\label{tab:metrics}
	\end{table}
	
	\begin{table}[!ht] \footnotesize 
		\centering
		\caption{Average accuracy per category.}
		\begin{tabular}{lr}
			\toprule
			\textbf{Category (Images)} & Accuracy \\
			\midrule
			\textbf{Category 1 (73)} & 0.7585 x 73 = 55.3709 \\
			\textbf{Category 2 (220)} & 0.7471 x 220 = 164.3572 \\
			\textbf{Category 3 (45)} & 0.7682 x 45 = 34.5702 \\
			\textbf{Category 4 (140)} & 0.7909 x 140 = 110.7238 \\
			\textbf{Category 5 (120)} & 0.7506 x 120 = 90.0745 \\
			\textbf{Category 6 (170)} & 0.8295 x 170 = 141.0067 \\
			\textbf{Category 7 (115)} & 0.7401 x 115 = 85.1139 \\
			\textbf{Category 8 (457)} & 0.8351 x 457 = 381.6530 \\
			\textbf{Category 9 (45)} & 0.7646 x 45 = 34.4083 \\
			\textbf{Category 10 (115)} & 0.8005 x 115 = 92.0603 \\
			\midrule
			\multicolumn{1}{r}{\textbf{SUM}} & \multicolumn{1}{r}{\textbf{1189.34}} \\
			\midrule
			\multicolumn{1}{r}{\textbf{Overall accuracy}} & \multicolumn{1}{r}{\textbf{0.7929}} \\
			\bottomrule
			{\tiny Overall accuracy = SUM/1,500} &  \\
		\end{tabular}
		\label{tab:exemplo_calculo_metricas}
	\end{table}
	
	\begin{figure}[!ht] \footnotesize 
		\begin{center}
			\includegraphics[scale=0.5]{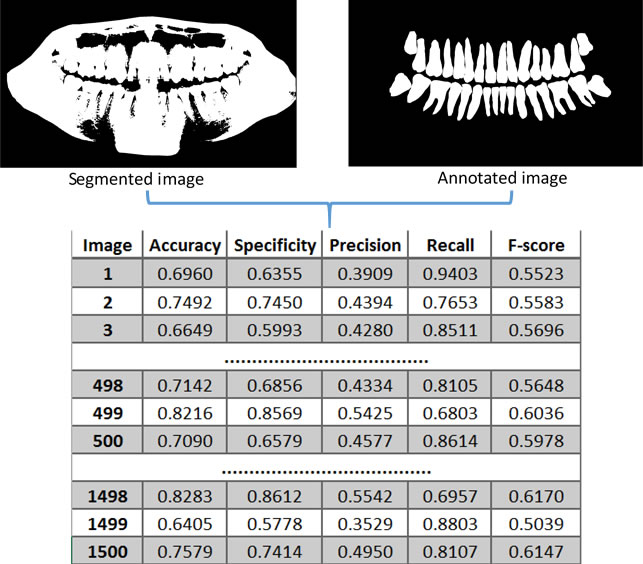}
			\caption{Obtaining the values of the metrics used over each image of the data set.} \label{figura:resultados_analise_imagem}
		\end{center}
	\end{figure}
	
	\subsection{Performance analysis of the segmentation methods} \label{sec_Segmentation_methods_parsed}
	
	The following metrics were used to evaluate the segmentation methods studied: Accuracy, specificity, precision, recall (sensitivity) and F-score, which are commonly used in the field of computer vision for performance analysis of segmentation. Table \ref{tab:metrics} presents a summary of these metrics.
	
	\begin{figure}[!ht] \footnotesize 
		\begin{center}
			\includegraphics[scale=0.5]{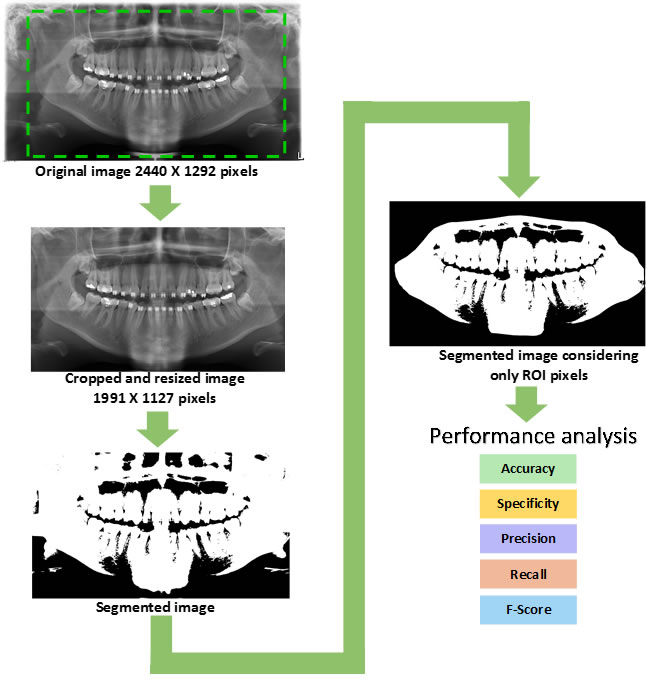}
			\caption{Steps of the performance analysis of segmentation algorithms.} \label{figura:metodologia_analise}
		\end{center}
	\end{figure}
	
	\begin{table}[!ht] \footnotesize   
		\centering
		\caption{Segmentation methods evaluated in the present study.}
		\begin{tabular}{p{2.7cm}|p{8.4cm}}
			\toprule
			\textbf{Category} & \textbf{Segmentation methods}
			\\
			\midrule
			Region-based & 1) Region growing; 2) Region splitting and merging
			\\
			Thresholding-based & 3) Basic global thresholding; 4) Niblack  
			\\
			Clustering-based & 5) Fuzzy C-means clustering
			\\
			Boundary-based & 6) Sobel; 7) Canny; 8)Active contour without edges; 9) Level-set  
			\\
			Watershed & 10) Marker-controlled watershed
			\\		
			\bottomrule
		\end{tabular}
		\label{tab:Segmentation_methods_evaluated}
	\end{table}
	
	\subsubsection{Methodology of the performance analysis}
	
	Only the image ROIs were considered to calculate the metrics for the evaluation of the segmentation methods. The process presented in Figure \ref{figura:resultados_analise_imagem} was carried out on all the segmented images obtained by each one of the 10 segmentation methods analyzed. Figure \ref{figura:metodologia_analise} illustrates the steps of the performance evaluation over 10 segmentation methods (see also Table \ref{tab:Segmentation_methods_evaluated} for a list of the evaluated methods). 
	
	\subsubsection{Computing the metrics over the data set}
	
	Table \ref{tab:exemplo_calculo_metricas} summarizes the process to calculate the accuracy for all images in each category, using only one segmentation method. Parameters of each method were optimized for best performance. On each category, the average accuracy was computed. After that, to find the accuracy for all images in the data set, the average accuracy was multiplied by the number of images in each category, obtaining a weighted sum for all images in the data set ($1,500$ images). By dividing by the number of images in the whole data set, we were able to find the average accuracy of the data set for one segmentation method. The same process was performed to calculate all other metrics (specificity, recall, precision and f-score), over each segmentation method. The segmentation methods evaluated in this work are summarized in Table \ref{tab:Segmentation_methods_evaluated}. Next, each one of the methods are discussed and evaluated.
	
	\paragraph{\textbf{1) Region growing}}
	
	Region growing is a method that groups pixels based on a predefined criterion to create larger regions. A standard approach for the region growing method is to perform calculations that generate sets of pixel values, whose properties group them close to the center (centroids) of the values we are looking for, so these values are used as seeds. Region growing needs two parameters to perform the segmentation operations, as follows:
	
	\begin{itemize}
		\item \textbf{Seeds} - Initial points to start the growth of regions. In the present work, the cells containing the $X$ and $Y$ coordinates of the centroids (center of the objects of interest) of the tooth regions (objects of interest in the images) were selected manually to serve as seeds. The seed values were grouped into vectors for each corresponding image, which served as the initial points for the execution of the region growing method;
		\item \textbf{\textit{Dist}} - Threshold that will work as a basis to indicate the similarity between the pixels that will be part of the region or not. The parameter \textit{dist} also corresponds to the conditional stop value of the algorithm. Thus, it is used to verify when the difference of the mean intensity between the pixels of a region and a new pixel becomes larger than the informed parameter, and therefore there are no more pixels to be inserted in the regions. The parameter \textit{dist} also corresponds to the conditional stop value of the algorithm. The best value found for the $dist$ parameter was 0.1.
		
	\end{itemize}
	
	\paragraph{\textbf{2) Region splitting and merging}}
	
	Segmentation based on division and union of regions is generally performed in four basic steps: 1) The image as a whole is considered as the area of initial interest; 2) The area of interest is examined to decide which of the pixels satisfy some criteria of similarity; 3) if true, the area of interest becomes part of a region in the image, receiving a label; 4) otherwise, the area of interest is divided and each one is successively considered as area of interest. After each division, a joining process is used to compare adjacent regions, putting them together if necessary. This process continues until no further division or no further union of regions are possible. The most granular level of division that can occur is when there are areas that contain only one pixel. Using this approach, all the regions that satisfy the similarity criterion are filled with 1's. Likewise, the regions that do not satisfy the similarity criterion are filled with 0's, thus creating a segmented image. The method needs two parameters:
	
	\begin{itemize}
		
		\item \textbf{\textit{qtdecomp}} - Minimum block size for decomposition (this parameter must be a positive integer), and set to 1 in our evaluation;
		\item \textbf{\textit{splitmerge}} - Similarity criterion used to indicate whether the region (block) should be divided or not. In the present work, we compared the standard deviation of the intensity of the pixels in the analyzed region. If the standard deviation is greater than the lowest intensity value of the pixels, then the region is divided.
		
	\end{itemize}
	
	\paragraph{\textbf{3) Basic global thresholding}}
	
	This method performs segmentation based on the histogram of the image. Assuming that \textbf{$f(x, y)$} corresponds to the histogram of an image, then to separate objects of interest from the background, an initial threshold ($T$) is chosen. Then any pixel of the image, represented by $(x, y)$, that is greater than $T$ is marked as an object of interest, otherwise the pixel is marked as the background. In our work, we used the following steps:
	
	\begin{enumerate}  
		\item{Estimate an initial value for the global limit, $T$ (we used the average pixel intensity of the ROI of each image analyzed);}
		
		\item{Segment the image through the threshold ($T$). Then, two groups of pixels appear: \(G_1 \), referring to pixels with values greater than $T$ and \(G_2 \), referring to pixels with values less than or equal to $T$;}
		
		\item{Calculate the mean intensity values, \(m_1\) and \(m_2\), of the pixels in \(G_1\) and \(G_2\), respectively;}  
		
		\item{Calculate $(m_1 + m_2)/2$ to obtain a new threshold ($T$) value;} 
		
		\item{Repeat steps 2 to 4 until the value of $T$, is less than a positive value predefined by a parameter \({\Delta} \)T. The larger the \({\Delta}\)T, the less interactions the method will perform. For the experiments, 0.5 was the best value found for \({\Delta}\)T, in our work;} 
		
		\item{Finally, converting the grayscale image into a binary image using the $\textbf{T / den}$ threshold, where $T$ is the threshold obtained in the previous steps and $den$ denote an integer value, which scales the maximum value of the ratio of $\textbf{T / den}$ to 1. The output image is binarized by replacing all pixels of the input image that are of intensity greater than the threshold $\textbf{T / den}$ by the value 1 and all other pixels by the value 0.} 
		
	\end{enumerate}
	
	\paragraph{\textbf{4) Niblack method}}
	
	Based on two measures of local statistics: mean and standard deviation within a neighborhood block of size n$\times$n, a threshold $T(x, y)$ for each pixel is calculated. Then, as the neighborhood block moves, it involves different neighborhoods, obtaining new thresholds, $T$, at each location $(x, y)$. Local standard deviation and mean are useful to determine local thresholds, because they are descriptors of contrast and luminosity. When the contrast or luminosity are found with great intensity in the images, they hinder the segmentation process in methods that use a single global threshold, such as the basic global thresholding. Adaptations of segmentation using local variable thresholding have been proposed in the literature. However, the method originally proposed in \citep{Niblack1985} was evaluated here. The local threshold is calculated with a block of size \textit{n$\times$n}, according to
	
	\begin{equation}
	T(x,y) = m(x,y) + k * \sigma(x,y) \, ,
	\end{equation}	
	where \textit{m(x,y)} and \textit{\({\sigma}\)(x,y)} represent the mean and local standard deviation of the local block of size \textit{n$ \times $n}, respectively. \textbf{k} is an imbalance constant (also called bias) that modifies the local value obtained by the local standard deviation. \textbf{\textit{k}} equal to 1 was adopted to avoid modifying the standard deviation locally calculated. For each pixel, the following process for all the images was performed:
	
	\begin{enumerate}
		
		\item Calculate mean and standard deviation within the local block $(x,y)$;
		
		\item Calculate the threshold $T(x,y)$; 
		
		\item If the value of the pixel is greater than $T(x,y)$, one is assigned.
		
	\end{enumerate}
	
	\paragraph{\textbf{5) Fuzzy C-means clustering}}
	
	Fuzzy C-Means starts from a set $X_k$ = \{$X_1, X_2, \\ 
	X_3, ..., X_n$\}  $\in$ $R^p$, where $X_k$ is a characteristic vector for all k $\in$ \{1, 2, ..., n\}, and $R^{p}$ is the p-dimensional space \citep{Bezdek1981}. A \textit{fuzzy} pseudopartition, denoted by P = \{$U_1, U_2,. . ., U_c$\}, must satisfy the following
	
	\begin{equation} \label{eq_cmenas_one}
	\sum_{i=1}^{c}U_1(X_k)=1 \, ,
	\end{equation}
	for every \textbf{k} $\in$ \{1, 2, ..., n\}, and \textbf{n} denotes the number of elements of the set $X$. The sum of the membership degrees of an element in all families must be equal to one. And,
	
	\begin{equation} \label{eq_cmenas_two}
	0<\sum_{k=1}^{n}U_i(X_k)<n \, ,
	\end{equation}	
	for every \textbf{i} $\in$  \{1, 2, ..., c\} and \textbf{c} represents the number of classes. Therefore, the sum of the membership degrees of all the elements of a family must be less than the number of elements in the set \textbf{X}. $U_i$ is the degree of relevance for $X_k$ in a cluster. The way to determine if the algorithm based on the Fuzzy C-means method finds an optimal fuzzy partition is defined by the objective function:
	
	\begin{equation} \label{eq_cmenas_three}
	J_m=\sum_{t=1}^{n}\sum_{c=1}^{k}U_{ci}^m\left \| X_{mn}-V_{mc} \right \|^2 \, ,
	\end{equation}	
	where $V_{mc} = (V_{11},...,V_{mc})$ is the matrix containing the centers of the clusters, $M$ is the fuzzy coefficient responsible for the degree of fuzing of the elements $X_{mn}$ and $V_{mc}$. The objective function is used to obtain the clusters by calculating the Euclidean distance between the image data and the cluster centers. The center $V_{mc}$ of each cluster $\textbf{c} (c = 1, ..., k)$ for an interaction $t$ is given by
	
	\begin{equation} \label{eq_cmenas_four}	
	V_{mc}^{(t)}=\frac{\sum_{i=1}^{n}(U_{ci}^{(t)})^MX_{mn}}{\sum_{i=1}^{n}(U_{ci}^{(t)})^M} \, .
	\end{equation}
	
	To perform the segmentation using the Fuzzy C-means method, the following parameters  were used: 
	
	\begin{itemize}
		
		\item Fuzzy coefficient $M$ equal to 2, responsible for the degree of fuzzification; 
		
		\item Stop criterion using a number of iterations (100), or if the method reaches the minimum error rate (0.00001).
		
	\end{itemize}
	
	The formation of clusters continues until the maximum number of iterations is completed, or when the minimum error rate is reached. 
	
	\paragraph{\textbf{6) Sobel}}
	Sobel works as an operator that calculates finite differences to identify the edges of the image. To perform the segmentation using the Sobel edge detector, we used an automatically threshold \textbf{T}, calculated from the image pixels.  
	
	\paragraph{\textbf{7) Canny}}
	
	In Canny \citep{Canny1986} detector, the edges are identified by observing the maximum location of the gradient of $f(x, y)$. Canny performs the following steps:
	
	\begin{enumerate}
		
		\item Smooth the Gaussian filter image first;
		
		\item The local gradient, $[g_x^2 + g_y^2]^{1/2}$, and the edge direction, $tan^{-1}(g_x/g_y)$, are computed at each point; 
		
		\item The edges are identified in step (2). Known the directions of the edge, a \textit{non-maximum suppression} is performed; this is done by tracing the border and suppressing pixel values (setting them to zero) that are not considered edge pixels. Two thresholds, T1 and T2, with T1 < T2 (automatically calculated based on each image);
		
		\item Finally, the edges detection of the image is performed considering the pixels that have values greater than T2.
		
	\end{enumerate}
	
	\paragraph{\textbf{8) Active contour without edges}}
	
	This method is a variation of the model originally proposed by Chan and Vese \citep{Chan2001}, and works differently from the classical edge detection methods. This is able to detect objects whose boundaries are not defined by the gradient. 
	
	To perform the segmentation using the active contour without edges method, the following parameters were used: 
	
	\begin{itemize}
		\item Initial mask (a matrix of 0's and 1's), where the mask corresponds to 75\% of the image to be segmented); 
		
		\item Total number of iterations (500, in our case);
		
		\item A stopping term equal to 0.1, in our work.
		
	\end{itemize}
	
	\paragraph{\textbf{9) Level set method}}
	
	This method is a variation of the active contour method. Level set method (LSM) was originally proposed by \cite{Osher1988}. LSM can perform numeric calculations of curves and surfaces without requiring predefined criteria. The objective of LSM is to represent the limit of an object using a level adjustment function, usually represented mathematically by the variable $ \alpha $. The curves of the objects are obtained by calculating the value of $ \gamma $ through the set of levels of $ \alpha $, given by:
	
	\begin{equation} \label{eq_level_set_method}
	\gamma=\{(x,y)|\alpha(x,y)=0\} \, , 
	\end{equation}
	and the $\alpha$ function has positive values within the region delimited by the $\gamma$ curve and negative values outside the curve.
    
	\paragraph{\textbf{10) Marker-controlled watershed}}
	
	The goal of the watershed segmentation method is to separate two adjacent regions, which present abrupt changing in the gradient values. Assuming that gradient values form a topographic surface with valleys and mountains, brighter pixels (e.g. teeth in X-ray images) correspond to those with the highest gradient, while the darker ones (e.g. valleys between teeth in X-ray images) would correspond to those with the lowest gradient. A variation of the watershed method is the marker-controlled watershed that prevents the occurrence of the phenomenon known as super-segmentation (excess of pixels that can not be attached to any other part of the image) using morphological operations of opening and closing to make adjustments to the gray levels of the image and avoid over-segmentation. 
	
	To segment the images in our data set by using the marker-controlled watershed method, the following steps are performed:
	
	\begin{enumerate}
		
		\item Compute a segmentation function, which tries to identify dark regions;
		
		\item Compute markers of the target object;
		
		\item Compute markers of the segments out of the target objects;
		
		\item Calculate the transformation of the segmentation function to obtain the position of the target objects and the positions of the background markers.
		
	\end{enumerate}
	
	\begin{table}[htbp] \footnotesize
		\centering
		\caption{Samples of the results of the segmentation methods evaluated (PART 1).}
		\begin{tabular}{C{1.9cm}C{1.95cm}C{1.95cm}C{1.95cm}C{1.95cm}C{1.9cm}}
			\toprule
			\multicolumn{1}{c}{\textbf{Method}} & \multicolumn{1}{c}{\textbf{Category 1}} & \multicolumn{1}{c}{\textbf{Category 2}} & \multicolumn{1}{c}{\textbf{Category 3}} & \multicolumn{1}{c}{\textbf{Category 4}} &
			\multicolumn{1}{c}{\textbf{Category 5}} \\
			\midrule
			\textbf{Region growing} & 	
			\includegraphics[scale=0.18]{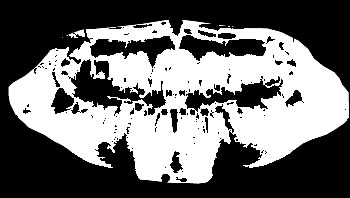} & \includegraphics[scale=0.18]{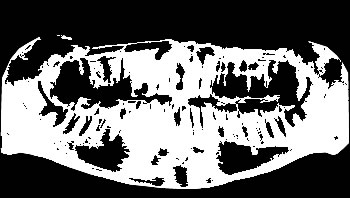} & 
			\includegraphics[scale=0.18]{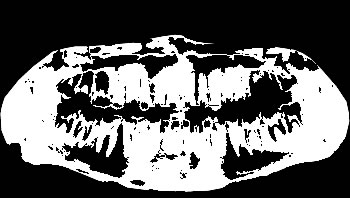} & 
			\includegraphics[scale=0.18]{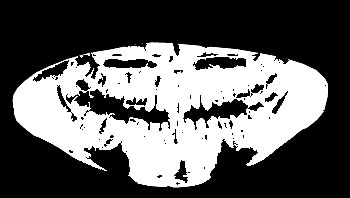} &
			\includegraphics[scale=0.18]{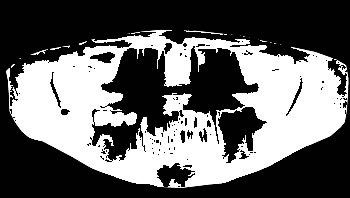} \\
			\textbf{Splitting and merging} & 
			\includegraphics[scale=0.18]{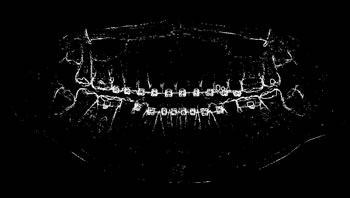} & \includegraphics[scale=0.18]{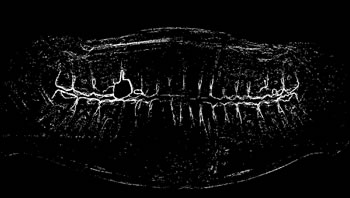} & 
			\includegraphics[scale=0.18]{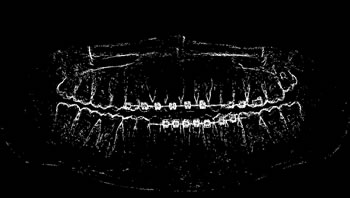} & 
			\includegraphics[scale=0.18]{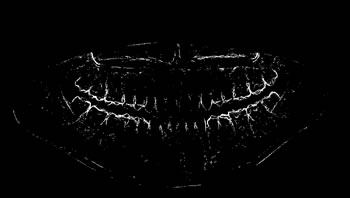} &
			\includegraphics[scale=0.18]{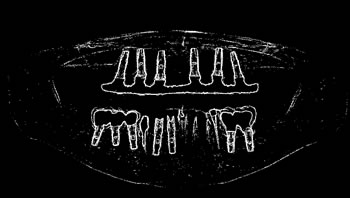} \\
			\textbf{Basic global threshold} & 
			\includegraphics[scale=0.18]{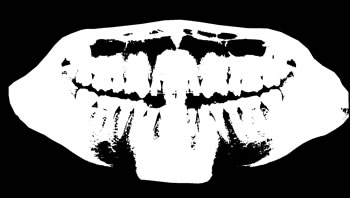} & \includegraphics[scale=0.18]{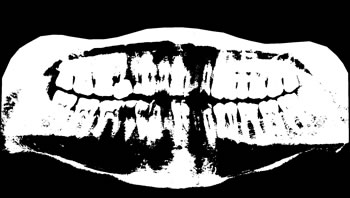} & 
			\includegraphics[scale=0.18]{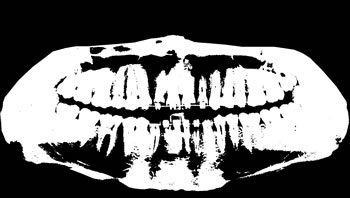} & 
			\includegraphics[scale=0.18]{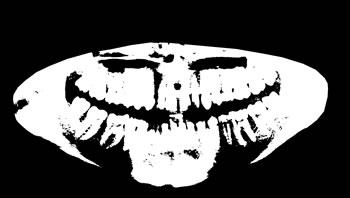} &
			\includegraphics[scale=0.18]{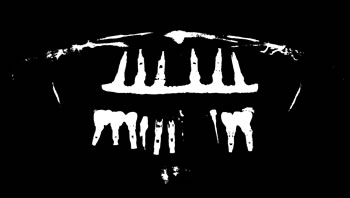} \\
			\textbf{Niblack method} & 
			\includegraphics[scale=0.18]{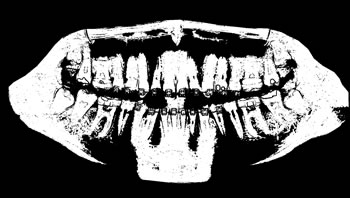} & \includegraphics[scale=0.18]{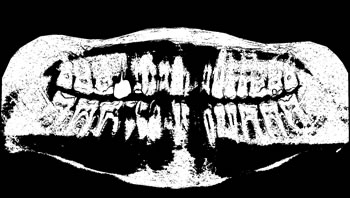} & 
			\includegraphics[scale=0.18]{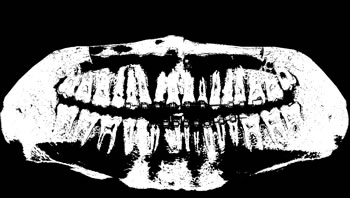} & 
			\includegraphics[scale=0.18]{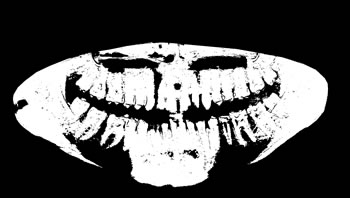} &
			\includegraphics[scale=0.18]{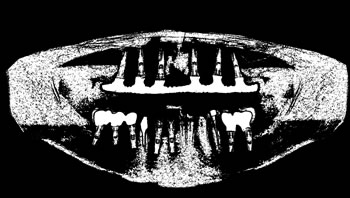} \\
			\textbf{Fuzzy c-means} & 
			\includegraphics[scale=0.18]{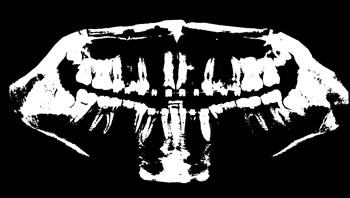} & \includegraphics[scale=0.18]{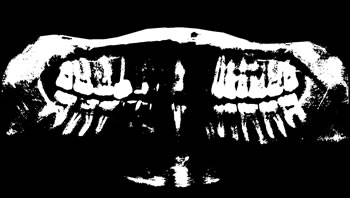} & 
			\includegraphics[scale=0.18]{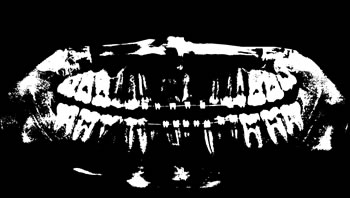} & 
			\includegraphics[scale=0.18]{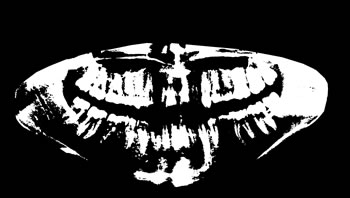} &
			\includegraphics[scale=0.18]{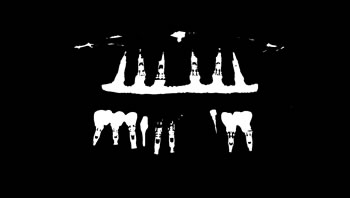} \\
			\textbf{Canny} & 
			\includegraphics[scale=0.18]{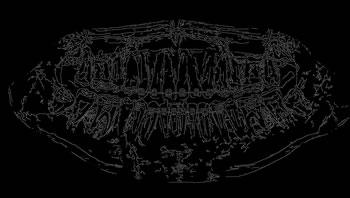} & \includegraphics[scale=0.18]{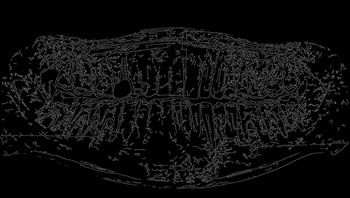} & 
			\includegraphics[scale=0.18]{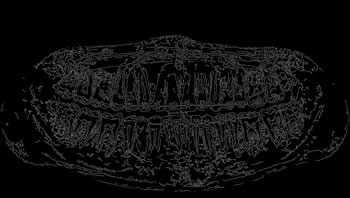} & 
			\includegraphics[scale=0.18]{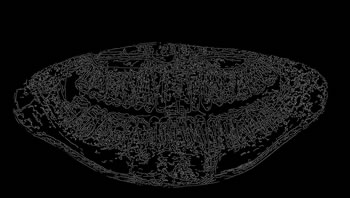} &
			\includegraphics[scale=0.18]{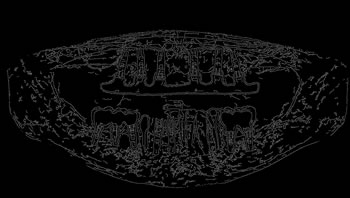} \\
			\textbf{Sobel} & 
			\includegraphics[scale=0.18]{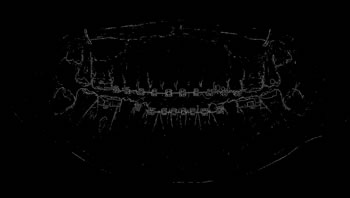} & \includegraphics[scale=0.18]{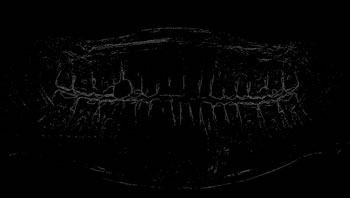} & 
			\includegraphics[scale=0.18]{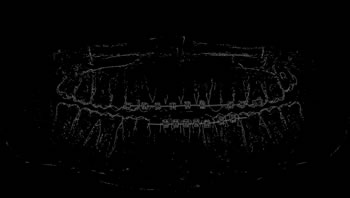} & 
			\includegraphics[scale=0.18]{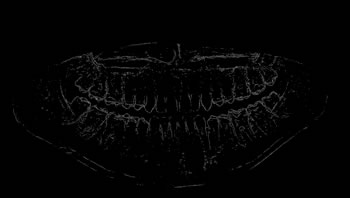} &
			\includegraphics[scale=0.18]{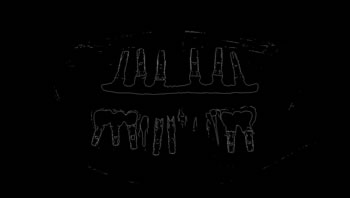} \\
			\textbf{Active contour without edges} & 
			\includegraphics[scale=0.18]{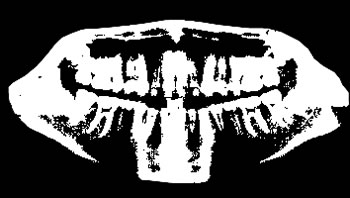} & \includegraphics[scale=0.18]{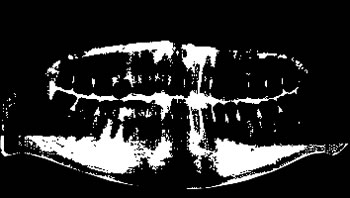} & 
			\includegraphics[scale=0.18]{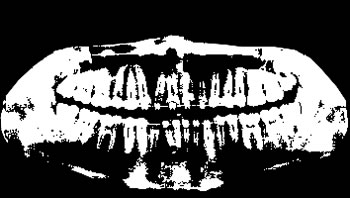} & 
			\includegraphics[scale=0.18]{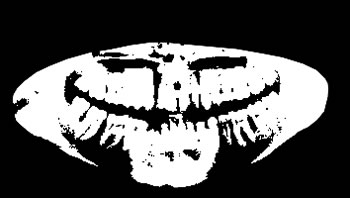} &
			\includegraphics[scale=0.18]{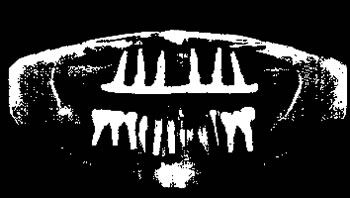} \\
			\textbf{Level set method} & 
			\includegraphics[scale=0.18]{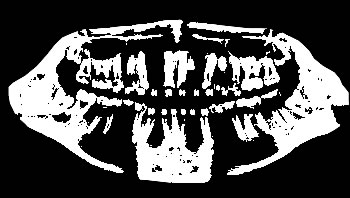} & \includegraphics[scale=0.18]{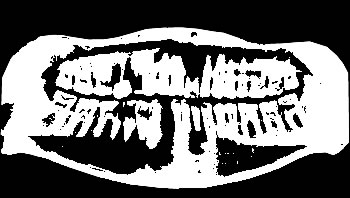} & 
			\includegraphics[scale=0.18]{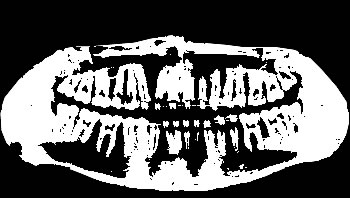} & 
			\includegraphics[scale=0.18]{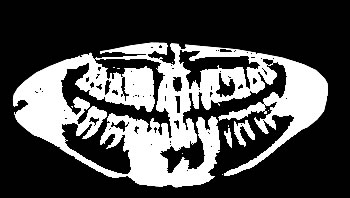} &
			\includegraphics[scale=0.18]{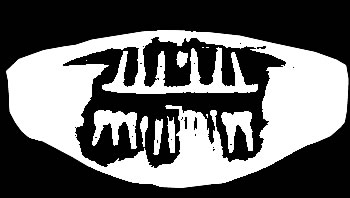} \\
			\textbf{Watershed} & 
			\includegraphics[scale=0.18]{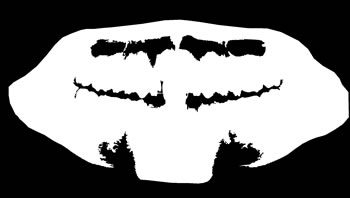} & \includegraphics[scale=0.18]{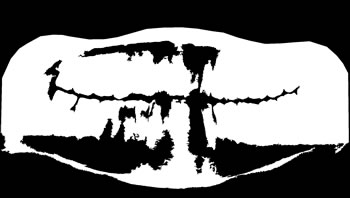} & 
			\includegraphics[scale=0.18]{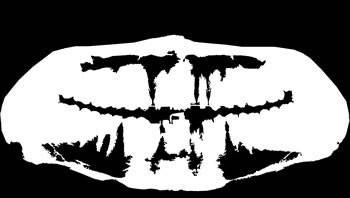} & 
			\includegraphics[scale=0.18]{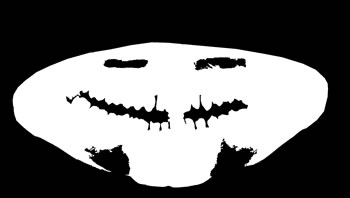} &
			\includegraphics[scale=0.18]{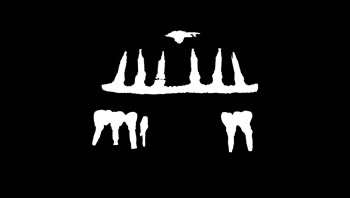} \\
			\bottomrule
		\end{tabular}
		\label{table:sample_results_part1}
	\end{table}
	
	\begin{table}[htbp] \footnotesize 
		\centering
		\caption{Samples of the results of the segmentation methods evaluated (PART 2).}
		\begin{tabular}{C{1.9cm}C{1.95cm}C{1.95cm}C{1.95cm}C{1.95cm}C{1.9cm}}
			\toprule
			\multicolumn{1}{c}{\textbf{Method}} & \multicolumn{1}{c}{\textbf{Category 6}} & \multicolumn{1}{c}{\textbf{Category 7}} & \multicolumn{1}{c}{\textbf{Category 8}} & \multicolumn{1}{c}{\textbf{Category 9}} &
			\multicolumn{1}{c}{\textbf{Category 10}} \\
			\midrule
			\textbf{Region growing} & 	
			\includegraphics[scale=0.18]{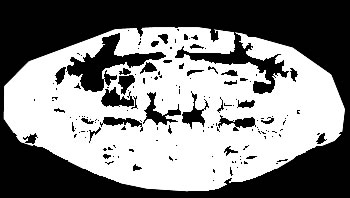} & \includegraphics[scale=0.18]{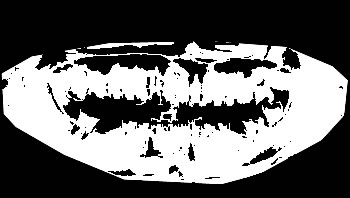} & 
			\includegraphics[scale=0.18]{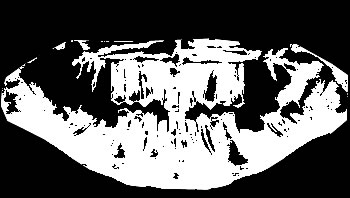} & 
			\includegraphics[scale=0.18]{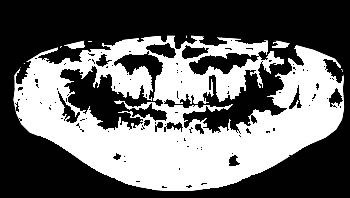} &
			\includegraphics[scale=0.18]{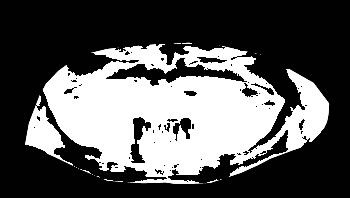} \\
			\textbf{Splitting and merging} & 
			\includegraphics[scale=0.18]{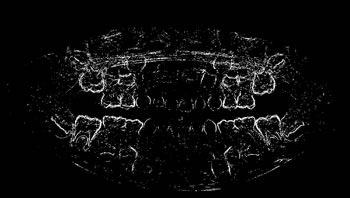} & \includegraphics[scale=0.18]{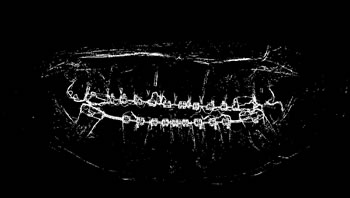} & 
			\includegraphics[scale=0.18]{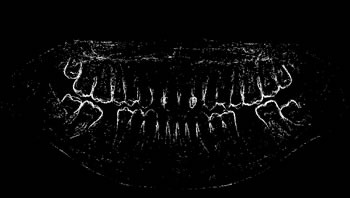} & 
			\includegraphics[scale=0.18]{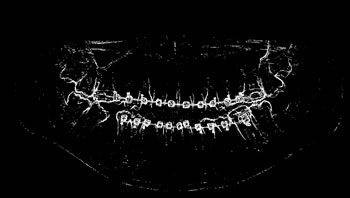} &
			\includegraphics[scale=0.18]{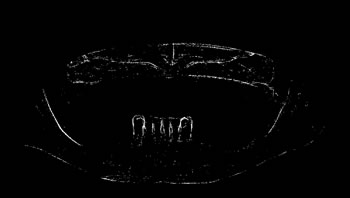} \\
			\textbf{Basic global threshold} & 
			\includegraphics[scale=0.18]{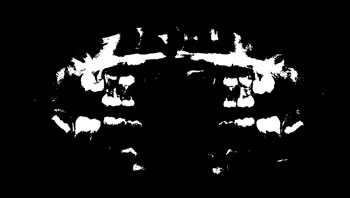} & \includegraphics[scale=0.18]{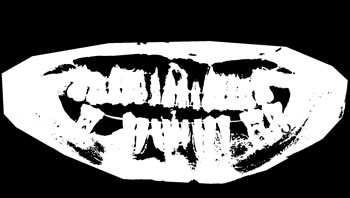} & 
			\includegraphics[scale=0.18]{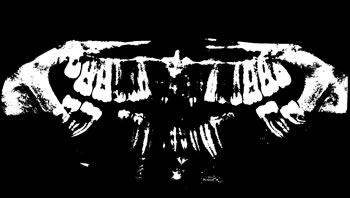} & 
			\includegraphics[scale=0.18]{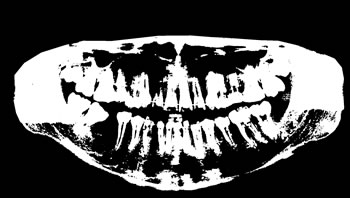} &
			\includegraphics[scale=0.18]{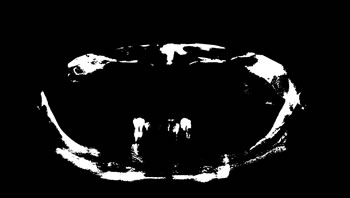} \\
			\textbf{Niblack method} & 
			\includegraphics[scale=0.18]{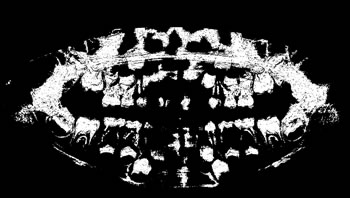} & \includegraphics[scale=0.18]{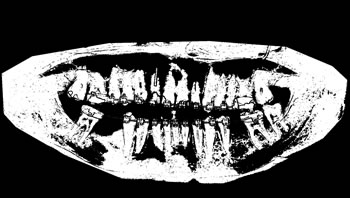} & 
			\includegraphics[scale=0.18]{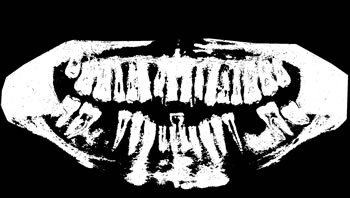} & 
			\includegraphics[scale=0.18]{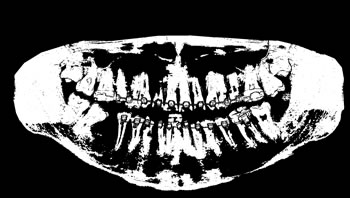} &
			\includegraphics[scale=0.18]{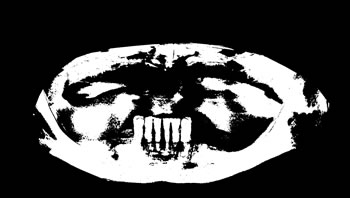} \\
			\textbf{Fuzzy c-means} & 
			\includegraphics[scale=0.18]{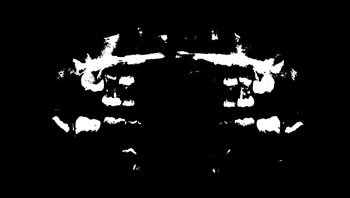} & \includegraphics[scale=0.18]{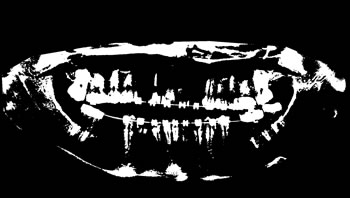} & 
			\includegraphics[scale=0.18]{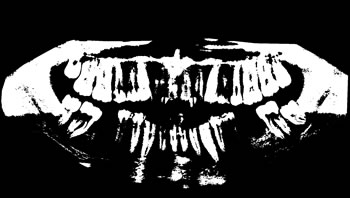} & 
			\includegraphics[scale=0.18]{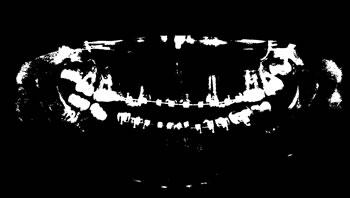} &
			\includegraphics[scale=0.18]{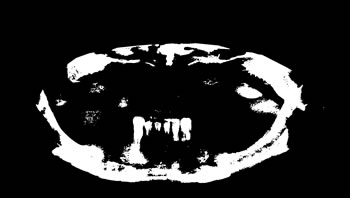} \\
			\textbf{Canny} & 
			\includegraphics[scale=0.18]{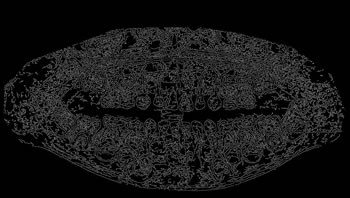} & \includegraphics[scale=0.18]{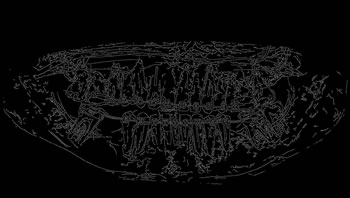} & 
			\includegraphics[scale=0.18]{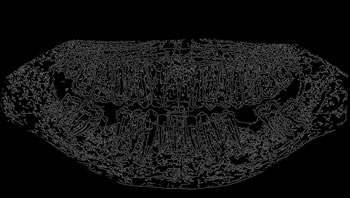} & 
			\includegraphics[scale=0.18]{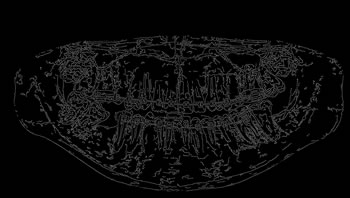} &
			\includegraphics[scale=0.18]{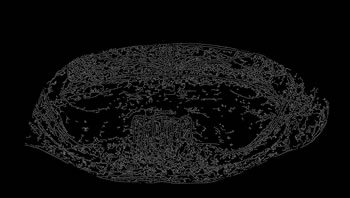} \\
			\textbf{Sobel} & 
			\includegraphics[scale=0.18]{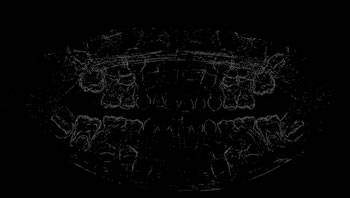} & \includegraphics[scale=0.18]{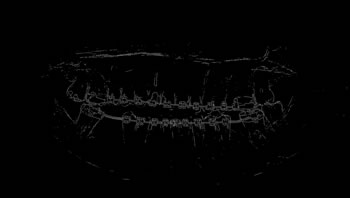} & 
			\includegraphics[scale=0.18]{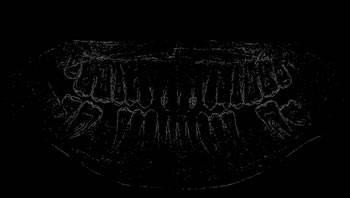} & 
			\includegraphics[scale=0.18]{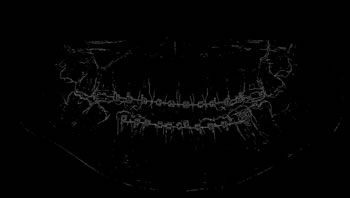} &
			\includegraphics[scale=0.18]{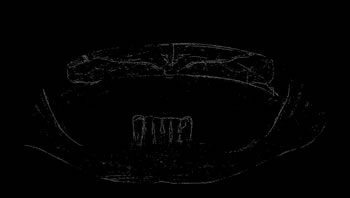} \\
			\textbf{Active contour without edges} & 
			\includegraphics[scale=0.18]{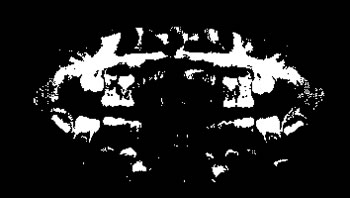} & \includegraphics[scale=0.18]{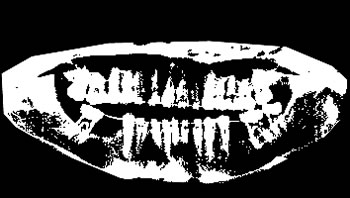} & 
			\includegraphics[scale=0.18]{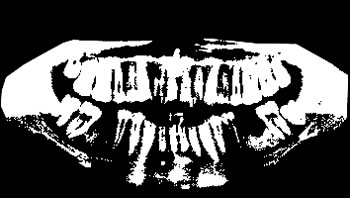} & 
			\includegraphics[scale=0.18]{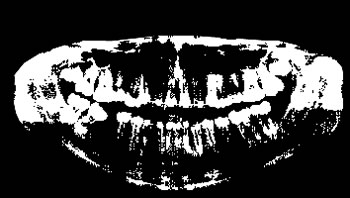} &
			\includegraphics[scale=0.18]{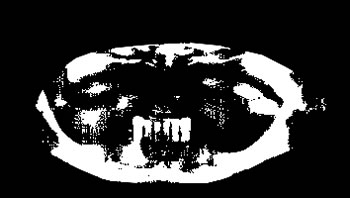} \\
			\textbf{Level set method} & 
			\includegraphics[scale=0.18]{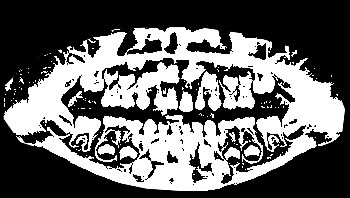} & \includegraphics[scale=0.18]{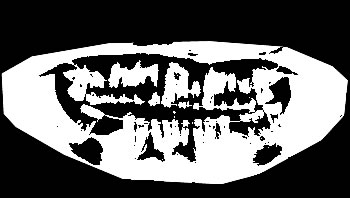} & 
			\includegraphics[scale=0.18]{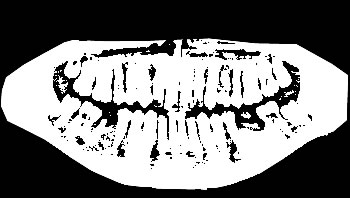} & 
			\includegraphics[scale=0.18]{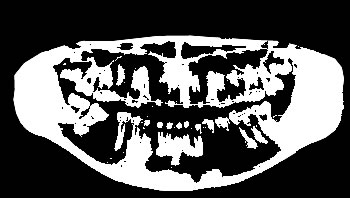} &
			\includegraphics[scale=0.18]{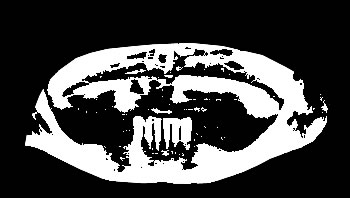} \\
			\textbf{Watershed} & 
			\includegraphics[scale=0.18]{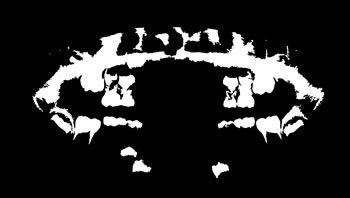} & \includegraphics[scale=0.18]{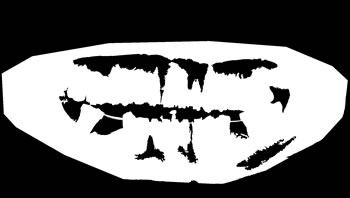} & 
			\includegraphics[scale=0.18]{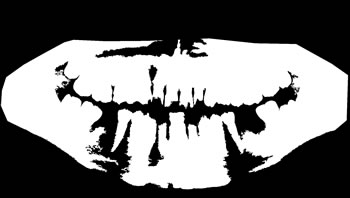} & 
			\includegraphics[scale=0.18]{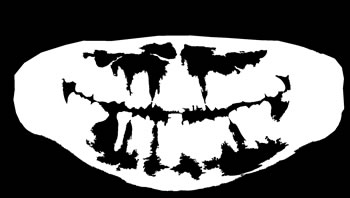} &
			\includegraphics[scale=0.18]{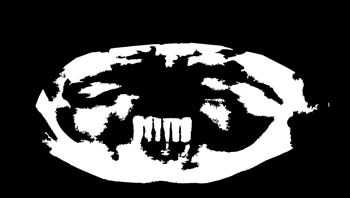} \\
			\bottomrule
		\end{tabular}
		\label{table:sample_results_part2}
	\end{table}
	
	\subsection{Result analysis} \label{sec:result_analysis}
	
	Tables \ref{table:sample_results_part1} and \ref{table:sample_results_part2} present samples of the results obtained by each of the segmentation methods evaluated, in each category. Table \ref{table:resultados_finais} summarizes the overall averages obtained by calculating the metrics, which were applied to evaluate the performance of each segmentation method. From Table \ref{table:resultados_finais}, we can observe that the splitting and merging-based and the sobel methods achieved almost perfect results in the specificity (which corresponds to the true negatives). This could be explained due to the characteristics of the two methods to focus on the edges. Similarly, the fuzzy C-means and canny methods also had over 90\% with respect to specificity. In contrast, these four segmentation methods obtained poor results in relation to the recall metric (which privileges the true positives in the evaluation). Thus, the images, segmented by algorithms based on splitting and merging, fuzzy C-means, canny and sobel, showed a predominance of true negatives in their results. However, when segmentation results in a predominance of elements of a class (for example, true negatives), it indicates that for a binary classification problem, the result of the segmentation algorithm is equivalent to the random suggestion of a class. Therefore, algorithms based on splitting and merging, fuzzy C-means, canny and sobel presented poor results when applied to the data set images used in the present work. 
	
	\begin{table}[!ht] \footnotesize  
		\centering
		\caption{Overall average results.}
		\begin{tabular}{L{3.1cm}ccccc}
			\toprule
			\multicolumn{1}{c}{\textbf{Method}} & \multicolumn{1}{c}{\textbf{Accuracy}} & \multicolumn{1}{c}{\textbf{Specificity}} & \multicolumn{1}{c}{\textbf{Precision}} & \multicolumn{1}{c}{\textbf{Recall}} & \multicolumn{1}{c}{\textbf{F-score}} \\
			\midrule
			\textbf{Region growing} & 0.6810  & 0.6948  & 0.3553 & 0.6341 & 0.4419 \\
			\textbf{Splitting/merging} & 0.8107 & \textbf{0.9958} & \textbf{0.8156} & 0.0807 & 0.1429 \\
			\textbf{Global thresholding} & 0.7929 & 0.8191 & 0.5202 & 0.6931 & 0.5621 \\
			\textbf{Niblack method} & 0.8182 & 0.8174 & 0.5129 & \textbf{0.8257} & \textbf{0.6138} \\
			\textbf{Fuzzy C-means} & \textbf{0.8234} & 0.9159 & 0.6186 & 0.4525 & 0.4939 \\
			\textbf{Canny} & 0.7927 & 0.9637 & 0.4502 & 0.1122 & 0.1751 \\
			\textbf{Sobel} & 0.8025 & 0.9954 & 0.6663 & 0.0360 & 0.0677 \\
			\textbf{Without edges} & 0.8020 & 0.8576 & 0.5111 & 0.5750 & 0.5209 \\
			\textbf{Level set method} & 0.7637 & 0.7842 & 0.4776 & 0.6808 & 0.5224 \\
			\textbf{Watershed} & 0.7658 & 0.7531 & 0.4782 & 0.8157 & 0.5782 \\
			\bottomrule
		\end{tabular}
		\label{table:resultados_finais}
	\end{table}
	
	The results presented in Table \ref{table:resultados_finais} also show that the niblack reached the highest value of the recall metric (approximately 83\%), indicating that the images segmented by niblack presented the highest number of true positives (pixels corresponding to the objects of interest in the analyzed images) and, therefore, few false negatives with respect to the other methods of segmentation evaluated. Niblack also obtained approximately 82\% in relation to the specificity metric, which corresponds to the true negatives that were correctly identified. Besides niblack, only the marker-controlled watershed method reached above 80\% of the recall metric. The marker-controlled watershed obtained lower results than the Nibliack method in all other analyzed metrics. The active contour without edges and the level set segmentation methods obtained less than 70\% of the recall metric; these methods also achieved poorer results when compared to the Niblack and marker-controlled watershed methods. 
    
    Considering the results, one can state that the segmentation process of panoramic X-ray images of the teeth based on thresholding, achieved a significant performance improvement when a local threshold (niblack method) was used, instead of using a single global threshold (basic global thresholding). Niblack was the one which presented superior performance to segment teeth.
	
	\section{Discussion and conclusions} \label{sec:discussion}
	
	From the images obtained with the X-ray, the dentist can analyze the entire dental structure and construct (if necessary) the patient's treatment plan. However, due to the lack of adequate automated resources to aid the analysis of dental X-ray images, the evaluation of these images occurs empirically, that is to say, only using the experience of the dentist. The difficulty of analyzing the dental X-ray images is still great when dealing with extra-oral radiographs, because these images are not restricted to only an isolated part of the teeth, as happens in the intra-oral images. In addition to the teeth, extra-oral X-rays also show the temporomandibular regions (jaw joints with the skull) and details originated by the bones of the nasal and facial areas. Other information on dental radiographs that make it difficult to analyze these images are: variations of patient-to-patient teeth, artifacts used for restorations and prostheses, poor image qualities caused by some conditions, such as noise, low contrast, homogeneity in regions that represent teeth and not teeth, space existing for a missing tooth, and limitation of acquisition methods, which sometimes result in unsuccessful segmentation. Therefore, the literature review carried out in the present study has revealed that there is still room to find an adequate method for segmentation of dental X-ray images that can be used as a basis for the construction of specialized systems to aid in dental diagnosis.
	
	It is noteworthy that the revised literature has been increasingly focused on research to segment dental X-ray images based on thresholding (see Section \ref{sec_review_segmentation_methods} and Table \ref{tab:Works_grouped_by_segmentation_methods}). Eighty percent (80\%) of the analyzed articles used intra-oral X-ray images in their experiments. Public data sets used in the works analyzed in our study indicate a preference in the literature for segmenting images that present only an isolated part of the teeth, rather than using extra-oral X-rays that show the entire dental structure of the mouth and face bones in a single image. The results of the present study also show that the majority of the papers reviewed (61\%) worked with data sets containing between 1 and 100 dental x-ray images (with only one work exploiting a data set with more than 500 images, containing 500 bitewing and 130 periapical dental radiographic films). This finding seems to indicate that most of the methods are not thoroughly evaluated. 
	
	One contribution of the present study was the construction of a data set with 1,500 panoramic X-ray images, characterizing a significant differential when compared to other works. Our data set has a diversity of images, with different characteristics and distributed in 10 categories that were defined. With our diverse and great-in-size data set, we performed an in-depth evaluation of the segmentation methods applied to the panoramic X-ray images of our dataset teeth using the following metrics: Accuracy, specificity, precision, recall, and F-score for performance measurement of the segmentation algorithms studied. After the performance evaluation of 10 segmentation methods over our proposed data set, we could conclude that none of the algorithms studied was able to completely isolate the objects of interest (teeth) in the data set images used in the present work, failing mainly because of the bone parts. 
	
\subsection{Future perspectives on exploiting learning-based segmentation methods}

	We can state that panoramic X-ray images of the teeth present characteristics that make the segmentation process difficult. It is realized that a possible way for segmentation of images in this branch can be the use of methods based on machine learning. Recently research has revealed interest in object recognition in images using segments \citep{Keypoints2004}, \citep{Arbel2012}, using techniques that combine segmentation and object recognition in images (also known as \textbf{semantic segmentation}). The work carried out in \citep{Carreira2012}, for example, presents how to explore grouping methods that calculate second order statistics in the form of symmetric matrices to combine image recognition and segmentation. The method proposes to efficiently perform second order statistical operations over a large number of regions of the image, finding clusters in shared areas of several overlapping regions. The role of image aggregation is to produce an overall description of a region of the image. Thus, a single descriptor can summarize local characteristics within a region, and can be used as input for a standard classifier. Most current clustering techniques calculate first-order statistics, for example, by performing the calculation of the maximum or average value of the characteristics extracted from a cluster \citep{Boureau2010}. The work proposed by \citep{Carreira2012} uses different types of second-order statistics, highlighting the local descriptors scale invariant feature transform (SIFT) \citep{Keypoints2004} and the Fisher coding \citep{Perronnin2010}, which also uses second order statistics to recognize objects in the images to perform the semantic segmentation. Second-order statistics could be exploited to segment dental X-ray images by learning the shape of the teeth while performing the segmentation.
	
	With no a priori information, finding segments of image objects is a remarkable skill of human vision. In the human visual system, when we look at an image, not all hypotheses are equally perceived by different people. For example, some people may recognize objects that are usually compact in their projection in the image and others may not be able to perceive these objects. In a computational view, the work performed in \cite{CarreiraSminchisescu2012} presents a new proposal to generate and classify hypotheses of objects in an image using processes from bottom-up and mid-level bands. The objects are segmented without prior knowledge about the their locations. The authors present how to classify the hypotheses of the objects through a model to predict the image segments, considering the properties of the region that composes the objects in the images studied. According to \cite{CarreiraSminchisescu2012}, real-world object statistics are not easy to identify in segmentation algorithms, sometimes leading to unsuccessful segmentation. One possibility to solve this problem would be to obtain the parameters of the segmentation algorithm, forming a model of machine learning using large amounts of annotated data. However, the local scope and inherently combinatory nature of the annotations of the images decrease the effectiveness of segmentation. While energy-minimization image segmentation problems generate multiple regions among objects, the work done in \citep{CarreiraSminchisescu2012} separated regions into individually connected components. Some characteristics harm the segmentation process: For example, in an image containing people hugged, segmenting people and their arms may require prior knowledge of the number of arms displayed in the image and the locations of the people may be in. Then, it is necessary to deal with such scenarios in an ascending way, that is to say, based on strong clues as continuity of the pixels that compose the objects of interest, and can be explored in the image analyzed. Still in the field of energy minimization, an observation that must be taken into account is the problem that leads to energy minimization that can be used to identify regions of image objects \citep{Tu2006}. In this sense, a promising direction that can be followed towards to improve the segmentation in dental X-ray images is the development of operations that can minimize energy functions to highlight the regions that represent objects of interest in the orthopantomography images. A big concern here is the overlap of image characteristics of the teeth with jaw and skull, in extra-oral X-rays images.
	
\subsection{Deep learning-based segmentation}

	The work found in \citep{Garcia-Garcia2017} provides a review of deep learning methods on semantic segmentation in various application areas. First, the authors describe the terminology of this field, as well as the mandatory background concepts. Then the key data sets and challenges are set out to help researchers on how to decide which ones best fit their needs and goals. The existing methods are reviewed, highlighting their contributions and their significance in the field. The authors conclude that semantic segmentation has been addressed with many success stories, although still remains an open problem. Given the recent success of deep learning methods in several areas of image pattern recognition, the use of these methods on a huge data set as ours could be a breakthrough in the field. As presented in \citep{Garcia-Garcia2017}, deep learning has proven to be extremely powerful to address the most diverse segmentation problems, so we can expect a flurry of innovation and lines of research over the next few years, exploiting studies that apply deep learning and semantic segmentation to dental X-ray images.

	
	\bibliography{mybibfile}

\begin{thebibliography}{}

\bibitem[Abaza et~al., 2009]{Abaza2009}
Abaza, A., Ross, A., and Ammar, H. (2009).
\newblock Retrieving dental radiographs for post-mortem identification.
\newblock In {\em Intl. Conf. on Image Processing}, pages 2537--2540.

\bibitem[Ajaz and Kathirvelu, 2013]{Ajaz2013}
Ajaz, A. and Kathirvelu, D. (2013).
\newblock {Dental biometrics: Computer aided human identification system using
  the dental panoramic radiographs}.
\newblock In {\em Intl. Conf. on Communication and Signal Processing}, pages
  717--721.

\bibitem[Ali et~al., 2015]{Ali2015}
Ali, R.~B., Ejbali, R., and Zaied, M. (2015).
\newblock {GPU-based Segmentation of Dental X-ray Images using Active Contours
  Without}.
\newblock In {\em Intl. Conf. on Intelligent Systems Design and Applications},
  volume~1, pages 505--510.

\bibitem[Alsmadi, 2015]{Alsmadi2015}
Alsmadi, M.~K. (2015).
\newblock A hybrid fuzzy c-means and neutrosophic for jaw lesions segmentation.
\newblock {\em Ain Shams Engineering Journal}.

\bibitem[Amer and Aqel, 2015]{Amer2015}
Amer, Y.~Y. and Aqel, M.~J. (2015).
\newblock An efficient segmentation algorithm for panoramic dental images.
\newblock {\em Procedia Computer Science}, 65:718--725.

\bibitem[An et~al., 2012]{An2012}
An, P.-l., Huang, P., Whe, P., Ang, H.~U., Science, C., Engineering, I., and
  Science, C. (2012).
\newblock {An automatic lesion detection method for dental x-ray images by
  segmentation using variational level set}.
\newblock In {\em Proceedings of the 2012 International Conference on Machine
  Learning and Cybernetics, Xian}, volume~1, pages 1821--1825.

\bibitem[Arbel et~al., 2012]{Arbel2012}
Arbel, P., Berkeley, B., Rd, W., and Park, M. (2012).
\newblock {Semantic Segmentation using Regions and Parts}.
\newblock pages 3378--3385.

\bibitem[Association, 1987]{AmericanDentalAssociation1987}
Association, A.~D. (1987).
\newblock Dental radiographic examinations - recommendations for patient
  selection and limiting radiation exposure.
\newblock {\em Journal of the American Medical Association},
  257(14):1929--1936.

\bibitem[Bezdek, 1981]{Bezdek1981}
Bezdek, J.~C. (1981).
\newblock Pattern recognition with fuzzy objective function algorithms.
\newblock {\em SIAM Review}, 25(3):442--442.

\bibitem[Boureau et~al., 2010]{Boureau2010}
Boureau, Y.-L., Ponce, J., and LeCun, Y. (2010).
\newblock {A Theoretical Analysis of Feature Pooling in Visual Recognition}.
\newblock {\em Icml}, pages 111--118.

\bibitem[Bruellmann et~al., 2016]{Bruellmann2016}
Bruellmann, D., Sander, S., and Schmidtmann, I. (2016).
\newblock The design of an fast fourier filter for enhancing diagnostically
  relevant structures -- endodontic files.
\newblock {\em Computers in Biology and Medicine}, 72:212--217.

\bibitem[Cameriere et~al., 2015]{Cameriere2015}
Cameriere, R., {De Luca}, S., Egidi, N., Bacaloni, M., Maponi, P., Ferrante,
  L., and Cingolani, M. (2015).
\newblock Automatic age estimation in adults by analysis of canine pulp/tooth
  ratio: Preliminary results.
\newblock {\em Journal of Forensic Radiology and Imaging}, 3(1):61--66.

\bibitem[Canny, 1986]{Canny1986}
Canny, J. (1986).
\newblock A computational approach to edge detection.
\newblock {\em IEEE Trans. on Patt. Analysis and Machine Intellig.},
  PAMI-8(6):679--698.

\bibitem[Carreira et~al., 2012]{Carreira2012}
Carreira, J., Caseiro, R., Batista, J., and Sminchisescu, C. (2012).
\newblock Semantic segmentation with second-order pooling.
\newblock {\em Lecture Notes in Computer Science}, 7578:430--443.

\bibitem[Carreira and Cristian, 2012]{CarreiraSminchisescu2012}
Carreira, J. and Cristian, S. (2012).
\newblock Cpmc: Automatic object segmentation using.
\newblock {\em IEEE Trans. on Pattern Analysis and Machine Intelligence},
  34(7):1312--1328.

\bibitem[Chan and Vese, 2001]{Chan2001}
Chan, T.~F. and Vese, L.~A. (2001).
\newblock {Active contours without edges}.
\newblock {\em IEEE Trans. on Image Processing}, 10(2):266--277.

\bibitem[Chen and Leung, 2004]{Chen2004}
Chen, S. and Leung, H. (2004).
\newblock Survey over image thresholding techniques and quantitative
  performance evaluation.
\newblock {\em Journal of Electronic Imaging}, 13(1):220.

\bibitem[Dighe and Revati, 2012]{Dighe2012}
Dighe, S. and Revati, S. (2012).
\newblock {Preprocessing, Segmentation and Matching of Dental Radiographs used
  in Dental Biometrics}.
\newblock 1(2278):52--56.

\bibitem[Economopoulos et~al., 2008]{Economopoulos2008}
Economopoulos, T., Matsopoulos, G.~K., Asvestas, P.~A., Gr{\"{o}}ndahl, K., and
  Gr{\"{o}}ndahl, H.~G. (2008).
\newblock Automatic correspondence using the enhanced hexagonal centre-based
  inner search algorithm for point-based dental image registration.
\newblock {\em Dentomaxillofacial Radiology}, 37(4):185--204.

\bibitem[Ehsani~Rad et~al., 2013]{EhsaniRad2013}
Ehsani~Rad, A., Shafry, M., Rahim, M., and Norouzi, A. (2013).
\newblock Digital dental x-ray image segmentation and feature extraction.
\newblock 11:3109--3114.

\bibitem[Garcia-Garcia et~al., 2017]{Garcia-Garcia2017}
Garcia-Garcia, A., Orts-Escolano, S., Oprea, S., Villena-Martinez, V., and
  Garcia-Rodriguez, J. (2017).
\newblock {A Review on Deep Learning Techniques Applied to Semantic
  Segmentation}.
\newblock pages 1--23.

\bibitem[Geraets et~al., 2007]{Geraets2007}
Geraets, W. G.~M., Verheij, J. G.~C., van~der Stelt, P.~F., Horner, K., Lindh,
  C., Nicopoulou-Karayianni, K., Jacobs, R., Harrison, E.~J., Adams, J.~E., and
  Devlin, H. (2007).
\newblock Prediction of bone mineral density with dental radiographs.
\newblock {\em Bone}, 40(5):1217--1221.

\bibitem[Gr{\'{a}}fov{\'{a}} et~al., 2013]{Grafova2013}
Gr{\'{a}}fov{\'{a}}, L., Ka{\v{s}}parov{\'{a}}, M., Kakawand, S.,
  Proch{\'{a}}zka, A., and Dost{\'{a}}lov{\'{a}}, T. (2013).
\newblock {Study of edge detection task in dental panoramic radiographs}.
\newblock {\em Dentomaxillofacial Radiology}, 42(7).

\bibitem[Hasan et~al., 2016]{Hassan2016}
Hasan, M.~M., Hassan, R., and Ismail, W. (2016).
\newblock Automatic segmentation of jaw from panoramic dental x-ray images
  using gvf snakes.
\newblock In {\em WAC}, volume~1, pages 1--6.

\bibitem[Huang and Hsu, 2008]{Huang2008}
Huang, C.~H. and Hsu, C.~Y. (2008).
\newblock {Computer-assisted orientation of dental periapical radiographs to
  the occlusal plane}.
\newblock {\em Oral Surgery, Oral Medicine, Oral Pathology, Oral Radiology and
  Endodontology}, 105(5):649--653.

\bibitem[Indraswari et~al., 2015]{Indraswari2015}
Indraswari, R., Arifin, A.~Z., Navastara, D.~A., and Jawas, N. (2015).
\newblock Teeth segmentation on dental panoramic radiographs using
  decimation-free directional filter banck thresholding and multistage adaptive
  thresholding.
\newblock In {\em Intl. Conf. on Information, Communication Technology and
  System}, volume~1, pages 49--54.

\bibitem[Jain and Chen, 2004]{Jain2004}
Jain, A.~K. and Chen, H. (2004).
\newblock {Matching of dental X-ray images for human identification}.
\newblock {\em Pattern Recognition}, 37(7):1519--1532.

\bibitem[Kaur and Kaur, 2016]{Kaur2016}
Kaur, J. and Kaur, J. (2016).
\newblock Dental image disease analysis using pso and backpropagation neural
  network classifier.
\newblock {\em Intl. Journal of Advanced Research in Computer Science and
  Software Engineering}, 6(4):158--160.

\bibitem[Keshtkar and Gueaieb, 2007]{Keshtkar2007}
Keshtkar, F. and Gueaieb, W. (2007).
\newblock {Segmentation of dental radiographs using a swarm intelligence
  approach}.
\newblock In {\em Canadian Conference on Electrical and Computer Engineering},
  pages 328--331.

\bibitem[Keypoints and Lowe, 2004]{Keypoints2004}
Keypoints, S.-i. and Lowe, D.~G. (2004).
\newblock Distinctive image features from.
\newblock {\em Intl. Journal of Computer Vision}, 60(2):91--110.

\bibitem[Li et~al., 2012]{Li2012}
Li, H., Sun, G., Sun, H., and Liu, W. (2012).
\newblock Watershed algorithm based on morphology for dental x-ray images
  segmentation.
\newblock In {\em Intl. Conference on Signal Processing Proceedings}, volume~2,
  pages 877--880.

\bibitem[Li et~al., 2007]{Li2007}
Li, S., Fevens, T., Krzyzak, A., Jin, C., and Li, S. (2007).
\newblock Semi-automatic computer aided lesion detection in dental x-rays using
  variational level set.
\newblock {\em Pattern Recognition}, 40(10):2861--2873.

\bibitem[Li et~al., 2006]{Li2006}
Li, S., Fevens, T., Krzyzak, A., and Li, S. (2006).
\newblock An automatic variational level set segmentation framework for
  computer aided dental x-rays analysis in clinical environments.
\newblock {\em Computerized Medical Imag. and Graph.}, 30(2):65--74.

\bibitem[Lin et~al., 2015]{Lin2015}
Lin, P.~L., Huang, P.~W., Huang, P.~Y., and Hsu, H.~C. (2015).
\newblock Alveolar bone-loss area localization in periodontitis radiographs
  based on threshold segmentation with a hybrid feature fused of intensity and
  the h-value of fractional brownian motion model.
\newblock {\em Computer Methods and Programs in Biomedicine}, 121(3):117--126.

\bibitem[Lin et~al., 2013]{Lin2013}
Lin, P.~L., Huang, P.~Y., and Huang, P.~W. (2013).
\newblock An effective teeth segmentation method for dental periapical
  radiographs based on local singularity.
\newblock In {\em Intl. Conf. on System Science and Engineering}, volume~1,
  pages 407--411.

\bibitem[Lin et~al., 2014]{Lin2014}
Lin, P.~L., Huang, P.~Y., Huang, P.~W., Hsu, H.~C., and Chen, C.~C. (2014).
\newblock {Teeth segmentation of dental periapical radiographs based on local
  singularity analysis}.
\newblock {\em Computer Methods and Programs in Biomedicine}, 113(2):433--445.

\bibitem[Lin et~al., 2010]{Lin2010}
Lin, P.~L., Lai, Y.~H., and Huang, P.~W. (2010).
\newblock {An effective classification and numbering system for dental bitewing
  radiographs using teeth region and contour information}.
\newblock {\em Pattern Recognition}, 43(4):1380--1392.

\bibitem[Lin et~al., 2012]{Lin2012}
Lin, P.~L., Lai, Y.~H., and Huang, P.~W. (2012).
\newblock {Dental biometrics: Human identification based on teeth and dental
  works in bitewing radiographs}.
\newblock {\em Patt. Recog.}, 45(3):934--946.

\bibitem[Lurie et~al., 2012]{Lurie2012}
Lurie, A., Tosoni, G.~M., Tsimikas, J., and Fitz, W. (2012).
\newblock {Recursive hierarchic segmentation analysis of bone mineral density
  changes on digital panoramic images}.
\newblock {\em Oral Surgery, Oral Medicine, Oral Pathology and Oral Radiology},
  113(4):549--558.

\bibitem[Modi and Desai, 2011]{Modi2011}
Modi, C.~K. and Desai, N.~P. (2011).
\newblock {A simple and novel algorithm for automatic selection of ROI for
  dental radiograph segmentation}.
\newblock In {\em Canadian Conference on Electrical and Computer Engineering},
  pages 000504--000507.

\bibitem[{Mohamed Razali} et~al., 2014]{MohamedRazali2014}
{Mohamed Razali}, M.~R., Ahmad, N.~S., {Mohd Zaki}, Z., and Ismail, W. (2014).
\newblock {Region of adaptive threshold segmentation between mean, median and
  otsu threshold for dental age assessment}.
\newblock In {\em Intl. Conf. on Computer, Communications, and Control
  Technology, Proceedings}, pages 353--356.

\bibitem[Niblack, 1985]{Niblack1985}
Niblack, W. (1985).
\newblock {\em An introduction to digital image processing}.
\newblock 1 edition.

\bibitem[Niroshika et~al., 2013]{Niroshika2013}
Niroshika, U. A.~A., Meegama, R. G.~N., and Fernando, T. G.~I. (2013).
\newblock {Active contour model to extract boundaries of teeth in dental X-ray
  images}.
\newblock In {\em Proceedings of the 8th International Conference on Computer
  Science and Education, ICCSE 2013}, pages 396--401.

\bibitem[Nomir and Abdel-Mottaleb, 2008a]{Nomir2008}
Nomir, O. and Abdel-Mottaleb, M. (2008a).
\newblock Fusion of matching algorithms for human identification using dental
  x-ray radiographs.
\newblock {\em IEEE Transactions on Inform. Forensics and Security},
  3(2):223--233.

\bibitem[Nomir and Abdel-Mottaleb, 2008b]{Nomir2008a}
Nomir, O. and Abdel-Mottaleb, M. (2008b).
\newblock Hierarchical contour matching for dental x-ray radiographs.
\newblock {\em Pattern Recognition}, 41(1):130--138.

\bibitem[Oliveira and Proen{\c{c}}a, 2011]{Oliveira2011}
Oliveira, J. and Proen{\c{c}}a, H. (2011).
\newblock {\em Caries Detection in Panoramic Dental X-ray Images}, pages
  175--190.
\newblock Springer Netherlands, Dordrecht.

\bibitem[Osher, 1988]{Osher1988}
Osher, S.~J. (1988).
\newblock Fronts propagating with curvature dependent speed.
\newblock {\em Computational Physics}, 79(1):1--5.

\bibitem[Paewinsky et~al., 2005]{Paewinsky2005}
Paewinsky, E., Pfeiffer, H., and Brinkmann, B. (2005).
\newblock Quantification of secondary dentine formation from orthopantomograms
  -- a contribution to forensic age estimation methods in adults.
\newblock {\em Intl. Journal of Legal Medicine}, 119(1):27--30.

\bibitem[Perronnin et~al., 2010]{Perronnin2010}
Perronnin, F., Jorge, S., Mensink, T., Perronnin, F., Jorge, S., Mensink, T.,
  Kernel, F., Perronnin, F., and Jorge, S. (2010).
\newblock Improving the fisher kernel for large-scale image classification to
  cite this version : Improving the fisher kernel for large-scale image
  classification.
\newblock In {\em Springer-Verlag Berlin Heidelberg}, pages 143--156.

\bibitem[Quinn and Sigl, 1980]{Quinn1980}
Quinn, R.~A. and Sigl, C.~C. (1980).
\newblock {\em Radiography in Modern Industry}.

\bibitem[Razali et~al., 2015]{Razali2015}
Razali, M. R.~M., Ahmad, N.~S., Hassan, R., Zaki, Z.~M., and Ismail, W. (2015).
\newblock Sobel and canny edges segmentations for the dental age assessment.
\newblock In {\em Intl. Conference on Computer Assisted System in Health},
  pages 62--66.

\bibitem[Said et~al., 2006]{Said2006}
Said, E., Nassar, D., Fahmy, G., and Ammar, H. (2006).
\newblock Teeth segmentation in digitized dental x-ray films using mathematical
  morphology.
\newblock {\em IEEE Transactions on Inform. Forensics and Security},
  1(2):178--189.

\bibitem[Senthilkumaran, 2012a]{Senthilkumaran2012a}
Senthilkumaran, N. (2012a).
\newblock Fuzzy logic approach to edge detection for dental x-ray image
  segmentation.
\newblock 3(5):5236--5238.

\bibitem[Senthilkumaran, 2012b]{Senthilkumaran2012}
Senthilkumaran, N. (2012b).
\newblock Genetic algorithm approach to edge detection for dental x-ray image
  segmentation.
\newblock {\em Intl. Journal of Advanced Research in Computer Science and
  Electronics Engineering}, 1(7):5236--5238.

\bibitem[Son and Tuan, 2016]{Son2016}
Son, L.~H. and Tuan, T.~M. (2016).
\newblock A cooperative semi-supervised fuzzy clustering framework for dental
  x-ray image segmentation.
\newblock {\em Expert Systems with Applications}, 46:380--393.

\bibitem[Tikhe et~al., 2016]{Tikhe2016}
Tikhe, S., Naik, A., Bhide, S., Saravanan, T., and Kaliyamurthie, K. (2016).
\newblock {Algorithm to Identify Enamel Caries and Interproximal Caries Using
  Dental Digital Radiographs}.
\newblock In {\em Intl. Advanced Computing Conference, IACC 2016}, pages
  225--228.

\bibitem[Trivedi et~al., 2015]{Trivedi2015}
Trivedi, D.~N., Kothari, A.~M., Shah, S., and Nikunj, S. (2015).
\newblock {Dental Image Matching By Canny Algorithm for Human Identification}.
\newblock {\em International Journal of Advanced Computer Research},
  4(17):985--990.

\bibitem[Tu and Zhu, 2006]{Tu2006}
Tu, Z. and Zhu, S.~C. (2006).
\newblock {Parsing images into regions, curves, and curve groups}.
\newblock {\em Intl. Journal of Computer Vision}, 69(2):223--249.

\bibitem[Wang et~al., 2016]{Wang2016}
Wang, C.~W., Huang, C.~T., Lee, J.~H., Li, C.~H., Chang, S.~W., Siao, M.~J.,
  Lai, T.~M., Ibragimov, B., Vrtovec, T., Ronneberger, O., Fischer, P., Cootes,
  T.~F., and Lindner, C. (2016).
\newblock {A benchmark for comparison of dental radiography analysis
  algorithms}.
\newblock {\em Medical Image Analysis}, 31:63--76.

\bibitem[Willems et~al., 2002]{Willems2002}
Willems, G., Moulin-Romsee, C., and Solheim, T. (2002).
\newblock Non-destructive dental-age calculation methods in adults: intra- and
  inter-observer effects.
\newblock {\em Forensic Science Intl.}, 126(3):221--226.

\end{thebibliography}
	
\end{document}